\documentclass{article} 
\usepackage{preprint,times}

\usepackage{amsmath,amsfonts,bm}

\def\eqref#1{equation~\ref{#1}}

\def\1{\bm{1}}

\DeclareMathAlphabet{\mathsfit}{\encodingdefault}{\sfdefault}{m}{sl}
\SetMathAlphabet{\mathsfit}{bold}{\encodingdefault}{\sfdefault}{bx}{n}

\definecolor{linkc}{rgb}{0, 0.44, 0.74}
\definecolor{eqc}{rgb}{1, 0, 0}
\usepackage[pagebackref=false,breaklinks=true,colorlinks=True,citecolor=linkc,linkcolor=eqc,bookmarks=false]{hyperref}
\usepackage{url}

\usepackage{multirow}
\usepackage{adjustbox}
\usepackage{booktabs}
\usepackage{enumitem}
\usepackage{xcolor}
\usepackage{array}
\usepackage{colortbl}
\usepackage{subcaption}
\usepackage{booktabs}
\usepackage{multirow}
\usepackage{siunitx}      
\usepackage{makecell}     
\usepackage{threeparttable} 
\usepackage{tcolorbox}
\usepackage{graphicx} 
\tcbuselibrary{breakable}
\usepackage{csquotes} 
\usepackage{tabularx}
\usepackage{wrapfig}  
\usepackage{adjustbox}
\usepackage{float}
\usepackage[normalem]{ulem}     
\title{\OURS}
\newcommand{\OURS}{VideoZoomer: Reinforcement-Learned Temporal Focusing for Long Video Reasoning}

\author{
Yang Ding$^{1}$\thanks{Equal contribution.}\ \ \ 
Yizhen Zhang$^{1}$$^{*}$\ \ \ 
Xin Lai$^{2}$\thanks{Project leader.}\ \ \ 
Ruihang Chu$^{1}$\ \ \ 
Yujiu Yang$^{1}\thanks{Corresponding authors.}$\\
$^1$Tsinghua University \quad $^2$The Chinese University of Hong Kong
}

\newcommand{\ours}{VideoZoomer}
\begin{document}

\maketitle

\begin{abstract}
Multimodal Large Language Models (MLLMs) have achieved remarkable progress in vision-language tasks yet remain limited in long video understanding due to the limited context window. Consequently, prevailing approaches tend to rely on uniform frame sampling or static pre-selection, which might overlook critical evidence and unable to correct its initial selection error during its reasoning process. To overcome these limitations, we propose \ours, a novel agentic framework that enables MLLMs to dynamically control their visual focus during reasoning. Starting from a coarse low-frame-rate overview, \ours~invokes a temporal zoom tool to obtain high-frame-rate clips at autonomously chosen moments, thereby progressively gathering fine-grained evidence in a multi-turn interactive manner. Accordingly, we adopt a two-stage training strategy: a cold-start supervised fine-tuning phase on a curated dataset of distilled exemplar and reflection trajectories, followed by reinforcement learning to further refine the agentic policy. Extensive experiments demonstrate that our 7B model delivers diverse and complex reasoning patterns, yielding strong performance across a broad set of long video understanding and reasoning benchmarks. These emergent capabilities allow it to consistently surpass existing open-source models and even rival proprietary systems on challenging tasks, while achieving superior efficiency under reduced frame budgets.
The code are avaliable at \url{https://github.com/zsgvivo/VideoZoomer}.
\end{abstract}

\section{Introduction}

With a clear task in mind, humans can efficiently navigate long and complex visual streams by dynamically allocating attention, selectively identifying salient events such as decisive actions in a sports match or key explanations in a lengthy lecture, while filtering out redundancy. This goal-directed ability underlies effective and efficient visual reasoning, as widely documented in cognitive science~\citep{Kietzmann2018DeepNN}, remains difficult to achieve in artificial intelligence. Although MLLMs perform strongly on image~\citep{qwen2_5vl,Chen2024ExpandingPB} and short-video tasks~\citep{Zhang2023VideoLLaMAAI}, they remain constrained in long-video comprehension tasks mainly due to their limited context window~\citep{gpt4o,gemini}.

The most common strategy to address this challenge is uniform frame sampling~\citep{llavanextvideo,llavavideo}, which selects frames at fixed intervals (e.g., two frames per second) to construct a subset that fits within context window. Nevertheless, this strategy is inherently limited, as it assumes all moments are equally important and further risks overlooking short but critical events while allocating context budget to redundant clip segments.
To address these limitations, prior work has investigated adaptive frame selection~\citep{yu2024frame,hu2025m,tang2025tspo}, where a lightweight selector module, conditioned on the text query, identifies salient frames before reasoning. 
While improving over uniform sampling, these methods are still inefficient because they are designed to select a fixed number of frames regradless of the problem's complexity. Second,  the design remains static and non-interactive. If the initial choice is suboptimal or misses key details, the model has no mechanism to correct the error or revisit the video. This fundamentally limits its performance on complex tasks that require iterative evidence gathering.

\begin{figure}[t]
    \centering
    \begin{subfigure}[b]{0.543\textwidth}
        \centering
        \includegraphics[width=\textwidth]{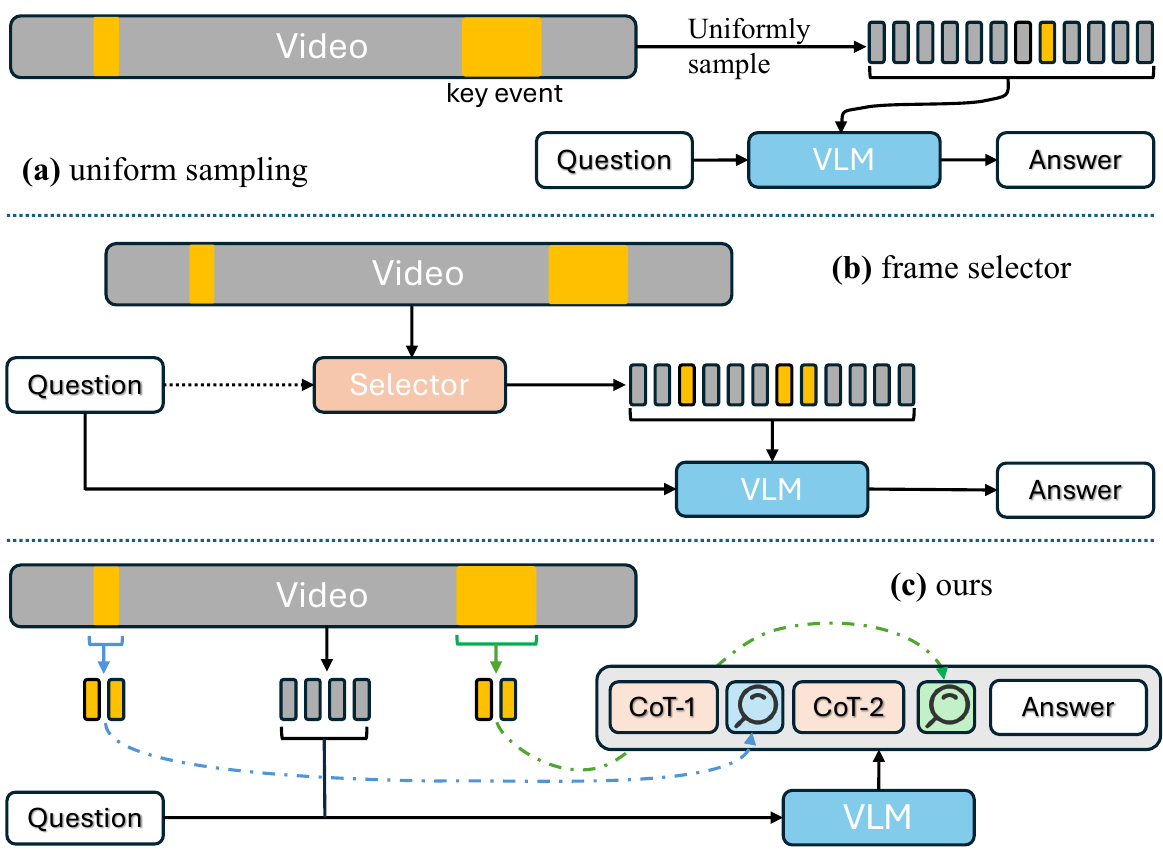} 
    \end{subfigure}
    \hfill
    \begin{subfigure}[b]{0.44\textwidth}
        \centering
        \includegraphics[width=\textwidth]{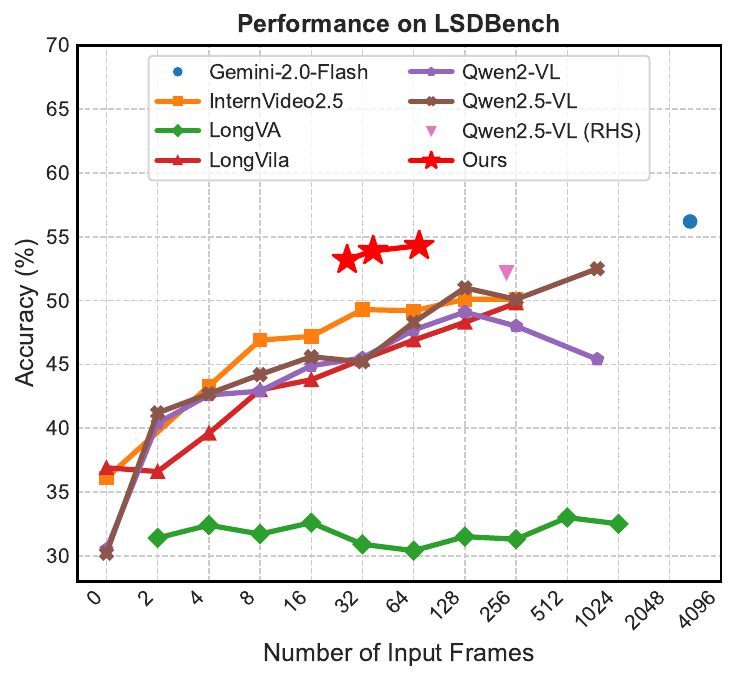}
    \end{subfigure}
    \caption{\textbf{Left}: Conceptual comparison of three long video reasoning frameworks: (a) uniform sampling, (b) with frame selector, and (c) \ours~(Ours). \textbf{Right}: Performance comparison of \ours~against various baseline models under different frame budgets on LSDBench.}
    \label{fig:teaser}
\end{figure}

To overcome the rigidity and inefficiency of prior methods, we propose \ours, a novel framework that empowers an MLLM to autonomously and dynamically control its visual focus during its reasoning process. As illustrated in Figure \ref{fig:teaser} (Left), instead of being a passive recipient of pre-selected frames, our model acts as an active agent. 

This yields two primary advantages: \textit{(i)} It is highly efficient: the agent begins with a coarse overview of low frame rates, only consuming a significant context budget when it decides to invoke a \texttt{<video\_zoom>} tool. This on-demand approach ensures that the model's context window is used dynamically and judiciously. \textit{(ii)} It is more performant: by learning a policy to request high-frame-rate clips of specific moments, the agent can correct initial oversights and gather detailed evidence precisely when and where it is needed. This dynamic, iterative evidence-gathering process avoids the critical information loss inherent in static methods and raises the upper bound on reasoning performance. Figure \ref{fig:teaser} (Right) demonstrates the practical benefit of this design on LSDBench~\citep{lsdbench}, a benchmark specifically designed to test a model's ability to find short, critical events in long videos. Our method achieves a better performance-efficiency trade-off, achieving superior accuracy compared to open source baselines while operating on a flexible and smaller frame budget.

Notably, training such an agent faces several challenges, a naive reinforcement learning approach would suffer from an inefficiently large action space and exhibit limited reasoning patterns. To address this, we introduce a two-stage training strategy. First, a cold-start Supervised Fine-Tuning (SFT) phase teaches the model the basics of tool using. Using a tailored dataset of exemplar trajectories, we train the model to understand the task format, master the syntax of tool calls, and develop a baseline reasoning capability. Crucially, to prevent the model from merely imitating a single, monotonous reasoning pattern, we enrich this dataset with reflection data, which exposes our model to more diverse and sophisticated problem-solving strategies. Second, with these foundational skills established, a Reinforcement Learning (RL) phase optimizes the model's tool interaction policy and reasoning capability, transforming it from a simple imitator into an adaptive agent that can generalize its strategy to unseen videos and questions.

We summarize our contributions as follows:
\begin{itemize}[nosep,leftmargin=12pt]
\item We propose \ours, a novel framework that reframes long video understanding as a sequential tool interaction task, enabling an MLLM to dynamically control its visual focus via multi-turn tool interaction.

\item We introduce a robust, two-stage training strategy: a cold-start phase using a tailored dataset of exemplar and reflection trajectories, followed by a reinforcement learning phase to optimize an efficient and effective agentic policy.

\item We demonstrate through extensive experiments that our model significantly outperforms existing open-source models on a wide range of long video understanding and reasoning benchmarks, in some cases even surpassing leading proprietary models with greater efficiency.
\end{itemize}

\section{Related Works}
\paragraph{Multimodal Reasoning Models.}
The remarkable success in LLMs~\citep{guo2025deepseek,team2025kimi,tan2025reason,jaech2024openaio1,yang2025qwen3} has demonstrated that reinforcement learning (RL) is a powerful paradigm for enhancing the complex reasoning capabilities.  Many works since then have tried to transfer this into multimodal domain. Methods such as MM-Eureka~\citep{meng2025mm} and VL-Rethinker~\citep{wang2025vl} have successfully adapted RL techniques to improve the vision-language reasoning abilities of MLLMs. 
More recently, Video-R1~\citep{videor1} further validated the efficacy of this approach specifically within the video domain.
Recently some works has tried further extend MLLMs with external tools like image cropping~\citep{zheng2025deepeyes,su2025pixel}, web search~\citep{wu2025mmsearch}, segmentation~\citep{liu2025seg}. However, most of these methods focus on image tasks and only interact with the environment for single turn, combining RL-driven reasoning and multi-turn tool use strategy for long video understanding is still underexplored.

\paragraph{Long Video Comprehension.}
Many works have tried to extend the ability of MLLMs in long video comprehension. A stream of research aims to reduce the number of visual tokens that need to be fed into the MLLM~\citep{liu2025video,yan2025crosslmm} through compression or selection modules.

A second, related approach focuses on selecting a sparse subset of the most salient frames from the entire video. Unlike uniform sampling, these methods aim to identify moments of high importance~\citep{tang2025tspo,hu2025m,wang2024vila}. While effective, the primary limitation of these methods is that the frame selection process is decoupled with its reasoning process, hindering it from learning more complex reasoning patterns. Methods like LongVILA-R1~\citep{longvilar1} focus on direct context extension by continuing training on long video datasets to handle longer video sequences.
Recently, a promising direction has emerged that leverages the powerful zero-shot capabilities of large proprietary models to act as agents. Frameworks like VideoDeepResearch~\citep{yuan2025videodeepresearch} and Deep Video Discovery~\citep{zhang2025deep} use prompting techniques to guide a strong LLM like Deepseek-R1~\citep{guo2025deepseek} or GPT 4.1~\citep{gpt4o} to iteratively explore a video with external tools. These training-free methods demonstrate the potential of agentic approach but rely on resource-intensive, closed-source models, making them difficult to optimize, reproduce, or deploy. In contrast, our work focuses on explicitly training a relatively small 7B open-source model to learn an efficient, agentic policy for long video comprehension.

\section{Method}

\subsection{Overview}
To address the challenge of efficient long video understanding with a constrained frame budget, we propose a novel framework, \ours, which empowers a large multimodal model to actively seek high-temporal-resolution information by invoking an external tool. Rather than relying on fixed or uniform sampling strategies, our model learns to dynamically and adaptively allocate its frame budget during its reasoning process. The core idea is to train an agent that learns an optimal policy for when and where to request high-frame-rate video clips, a process we call \enquote{temporal zoom-in}, to gather sufficient evidence for answering a given question. 

\begin{figure}[tb]
    \centering
    \includegraphics[width=1\linewidth]{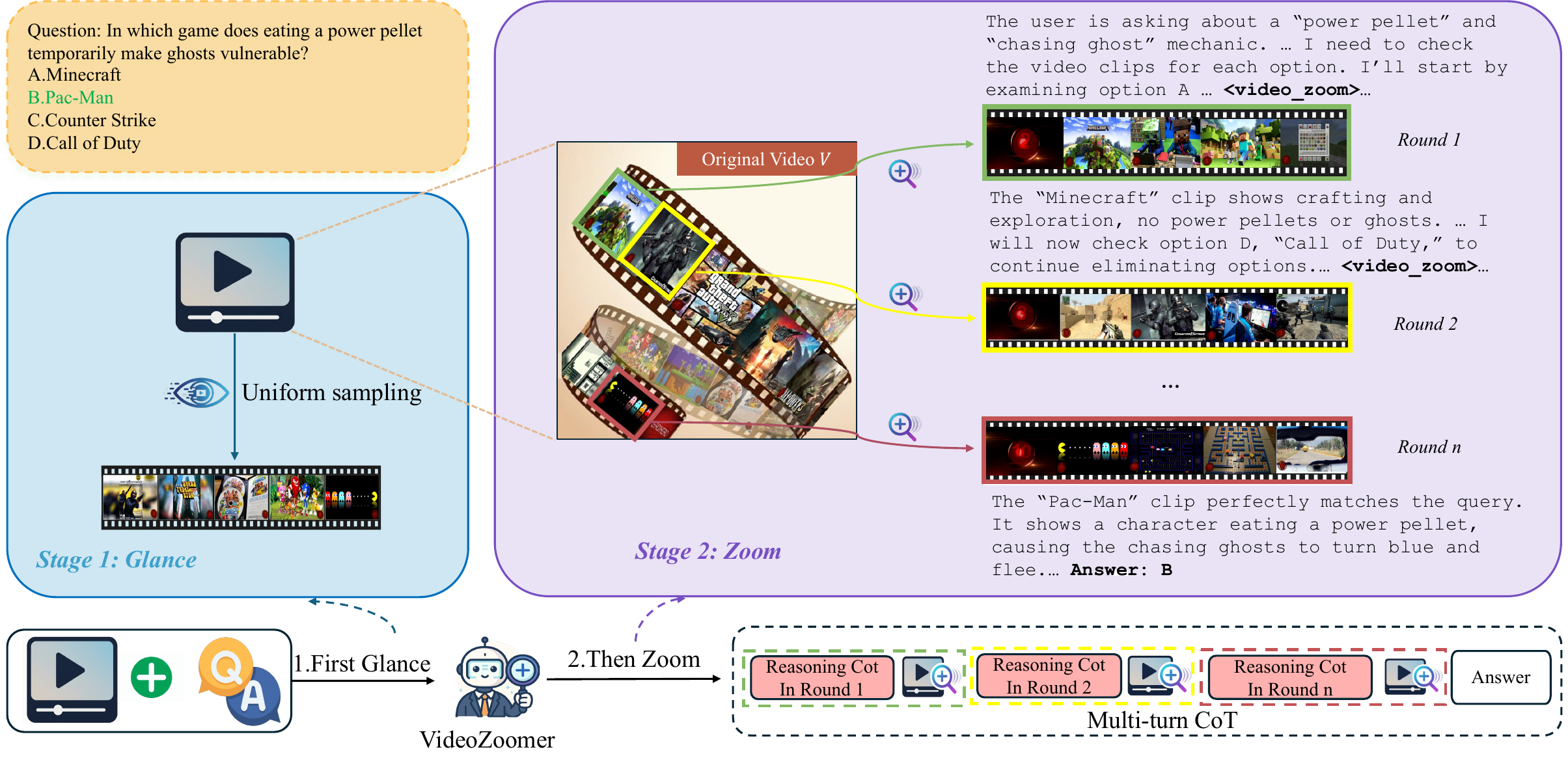}

    \caption{\textbf{\ours~framework for long video reasoning.} The process begins with a \enquote{Glance} where the model obtains a coarse overview of the video. It then enters an iterative \enquote{Zoom} phase, where it can invoke a \texttt{<video\_zoom>} tool to request high-fps clips and perform multi-turn reasoning. This process continues until the model procudes a final answer or reaches max turn limit. }
    \label{fig:method}
\end{figure}

\begin{figure}[tb]
    \centering
    \includegraphics[width=1.0\linewidth]{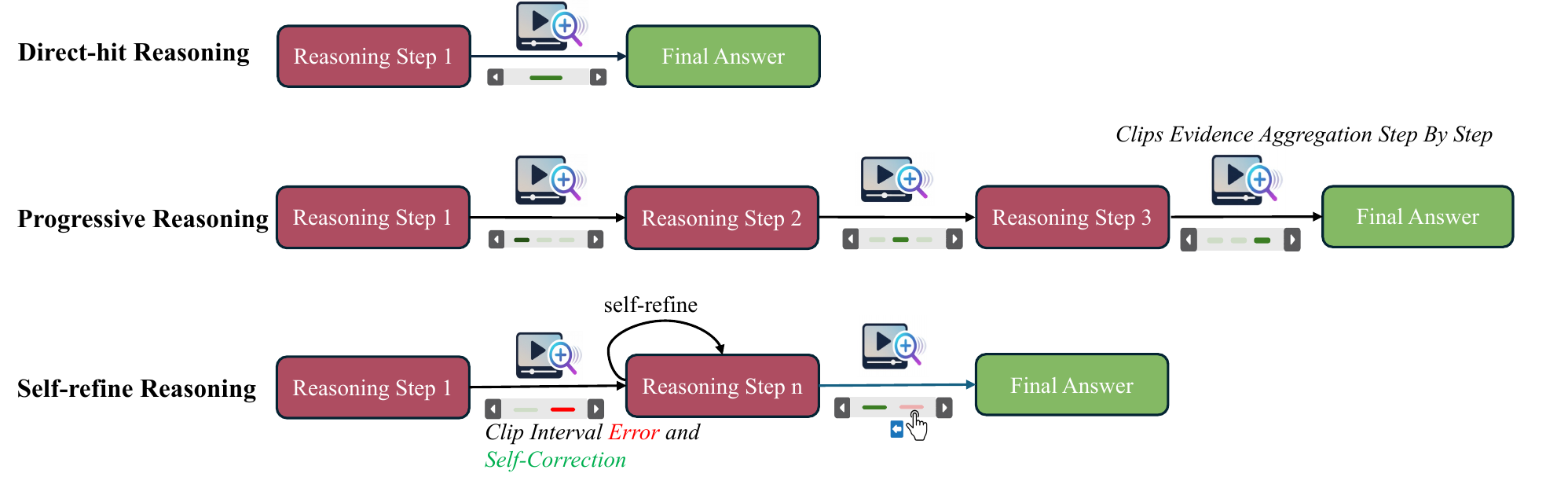}

    \caption{\textbf{Diverse reasoning patterns demonstrated by our model.} (a) Direct-hit Reasoning, (b) Progressive Reasoning, and (c) Self-refine Reasoning.}
    \label{fig:pattern}
    \vspace{-1em}
\end{figure}

As illustrated in Figure \ref{fig:method}, the \textcolor{blue}{strategy} is summarized as "first glance, then zoom": initially the model only has access to the query prompt $Q$ and a relatively low frame rate version of the video $V_{low}$, uniformly sampled as a default frame rate $f_{low}$, which provides a coarse, computationally inexpensive overview of the entire video. To answer the question accurately, especially when it pertains to fine-grained temporal events or rapid motions, the model may require more detailed visual information. We introduce a \texttt{<video\_zoom>} tool, which allows the model to request a specific time segment $[t_{start}, t_{end}]$ from the original video at a higher frame rate, $f_{high}$. Upon invoking this tool, the environment returns a high-resolution clip $V_{clip} = T(V, t_{start}, t_{end},f_{high})$. The agent's objective is to interact with the environment by iteratively calling the tool to gather visual evidence, this process continues until the agent determines it has sufficient information to produce a final answer. The agentic approach enables the model to develop diverse and complex reasoning strategies, as demonstrated in Figure \ref{fig:pattern}. Each tool calling is constrained by a frame budget $B$ (i.e. $f_{high}\times(t_{end}-t_{start})\leq B$), thus the total number of frames that can be requested from the high-resolution clips is limited by $B\times N$, where $N$ is the maximum number of interaction rounds. The environment returns an error message if the model makes an illegal request or exceeds the frame budget. The goal is to learn a policy $\pi$ that maximizes the quality of the final answer while adhering to the frame budget and tool call number constraints.

\subsection{Cold-start Initialization}
\label{coldstart}
\begin{figure}[t]
\centering
\includegraphics[width=0.95\linewidth]{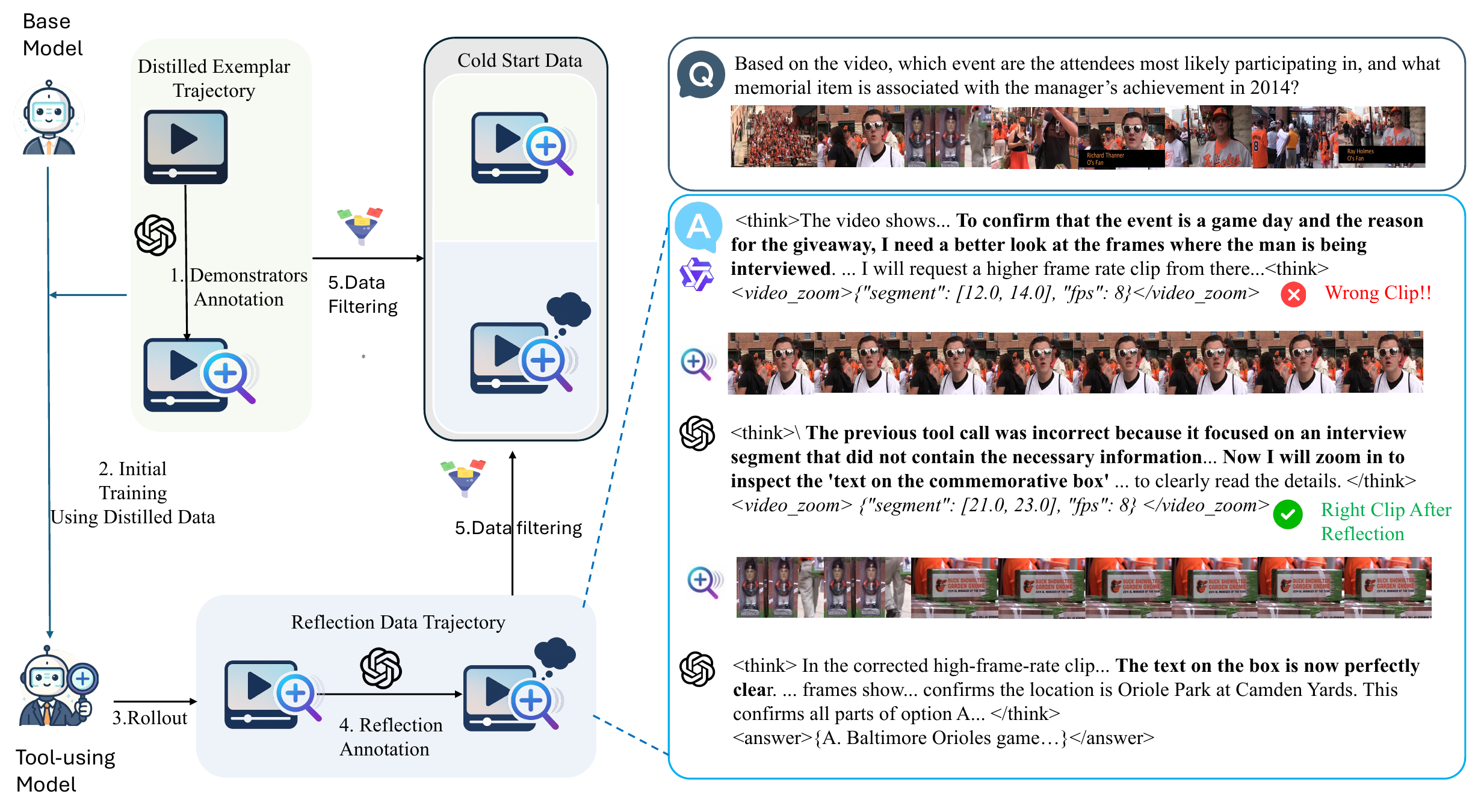}

\caption{\textbf{The pipeline for curating our cold-start dataset.} We first distill exemplar trajectories, then generate reflection data by having an expert model correct the failures of an initial agent. The final dataset combines both verified exemplar and reflection trajectories.}
\label{fig:sftpipe}
\vspace{-1em}
\end{figure}

Reinforcement learning from scratch on a complex, high-dimensional action space, such as generating structured tool calls, is notoriously sample-inefficient and prone to instability. To mitigate these challenges, we precede the RL phase with a supervised fine-tuning (SFT) stage designed to \enquote{cold-start} our agent. The primary objective of this stage is twofold: first, to equip a base multimodal model with the fundamental capability of understanding and invoking the \texttt{<video\_zoom>} tool in the correct format; and second, to expose it to a diverse range of reasoning patterns, which is crucial for effective exploration during subsequent RL training. To achieve this, we construct a specialized SFT dataset by curating high-quality, multi-turn interaction trajectories as illustrated in Figure \ref{fig:sftpipe}.

\paragraph{Distillation of Exemplar Trajectories.} The initial step is to generate a set of \enquote{golden} tool-use trajectories. We leverage state-of-the-art proprietary models, such as GPT-4o~\citep{gpt4o} and Gemini-2.5-pro~\citep{comanici2025gemini2.5}, as expert demonstrators. For each video-question pair in our training set, we prompt the expert model with the same system prompt and initial low-frame-rate video provided to our agent. The model then engages in a multi-turn interaction, iteratively calling the  \texttt{<video\_zoom>} tool until it gathers sufficient information to answer the question. This process yields a collection of complete trajectories, each containing the initial prompt, a sequence of tool calls, the corresponding high-frame-rate clip observations, and the final answer. These expert-generated trajectories serve as ideal examples of effective tool invocation and reasoning.

\paragraph{Augmentation with Reflection Data.} While SFT on only exemplar trajectories effectively teaches the model the format of tool use, we observed a significant limitation:  the resulting model tends to overfit the expert model's dominant reasoning patterns. This often leads to a \enquote{shallow} policy, where the model learns to call the tool at most once and then immediately outputs an answer, regardless of whether the retrieved clip was actually helpful or contained errors. This lack of perseverance and adaptability would severely hinder its ability to solve more complex problems requiring deeper, iterative reasoning. 

To address this and introduce more diverse and complex reasoning patterns, we generate reflection data. As shown in Figure \ref{fig:sftpipe}. The process begins by using our initial model trained only on exemplar data, to produce its own rollouts. We then identify trajectories where the model failed to answer correctly. These incorrect rollouts are subsequently fed back to the expert model, which is prompted to reflect on the flawed reasoning.  The model then identifies the mistake and generates a corrected, more robust reasoning path. This corrected path might involve additional tool calls or a different line of reasoning. This reflection process creates valuable training instances that explicitly teach the model how to recover from errors, critically evaluate the information returned by a tool, and when to persist with further investigation.
Furthermore, this on-policy-like data generation strategy ensures that the new trajectories are challenging yet achievable, mitigating distribution shift and stabilizing the transition from SFT to RL. 

The final cold-start dataset is a carefully curated combination of the distilled exemplar trajectories and reflection trajectories. Before inclusion, all candidate trajectories are passed through verifiers to ensure quality.  This resulted composite dataset, approximately 11,000 trajectories in total, provides a rich and balanced foundation for our base model.

\subsection{Multi-turn Tool-integrated Reinforcement Learning}

We employ GRPO~\citep{shao2024deepseekmath} for RL training due to its demonstrated efficacy in enhancing multimodal reasoning capabilities. 

We extend its original formula to multi-turn tool calling contexts and adapt a token-level loss mask on tool call trajectory, ignoring text and image tokens that are not generated by the model.

\paragraph{Reward Design.}
The design of the reward function is essential to guide the agent toward the desired behavior. Our reward is assigned at the end of each trajectory and is composed of three distinct components designed to promote accuracy, valid format, and exploration:
\begin{equation}
    R(x,y)=R_{acc}(x,y) + R_{\textcolor{blue}{format}}(y) + R_{tool}(y)
\end{equation}
The accuracy reward $R_{acc}$ is the primary task-oriented reward, it provides a strong positive signal if the agent's final answer is correct. 
The format reward $R_{format}$  validates the structure of the agent's response at each turn. This reward is set to a positive value if the model's output strictly adheres to the predefined format, and zero otherwise. Specifically, the agent receives a positive reward if every intermediate step correctly wraps its reasoning in \texttt{<think></think>} tags and be followed by either a valid \texttt{<video\_zoom></video\_zoom>} or a final answer enclosed in \texttt{<answer></answer>} tags.
A key challenge during early training is that a model unfamiliar with the \texttt{<video\_zoom>} tool may be hesitant to use it, often preferring to guess an answer directly. To solve this and encourage exploration, we introduce a bonus $R_{tool}$ for using the tool. To prevent the agent from learning to make redundant or unhelpful tool calls, this bonus is conditional: it is only awarded if the final answer is correct.

\section{Experiment}
\subsection{Experimental Setup}
\paragraph{Implementation Details.}
We initialize our model from Qwen-2.5-VL-7B-Instruct~\citep{qwen2_5vl} for its strong foundational capabilities and amenability to reinforcement learning. For cold-start initialization, we adapt the LLaMA-Factory~\citep{zheng2024llamafactory} framework. Our RL training and evaluation framework is based on verl~\citep{verl}, which we extended to support multi-turn tool-calling tasks and optimized for efficiency in long video training scenario.

For training data, we use LongVideoReason~\citep{longvilar1}, a long video QA dataset comprised of 52K high-quality question-reasoning-answer pairs.
In cold start stage, we trained our base model with a learning rate of $5\times 10^{-6}$ for 1 epoch on dataset we construct as described in Section \ref{coldstart}.
During RL stage, we use a learning rate of $1\times 10^{-6}$, rollout number of 16 and batchsize of 128. The model is initialized with 64 uniformly sampled frames. It can then perform up to 4 subsequent tool calls, each retrieving up to 16 frames of high-resolution clip from a segment of interest, before providing a final answer.
To improve training effectiveness and stability of RL training process, we also adapt clip-higher and dynamic sampling from DAPO~\citep{yu2025dapo}.
Further details are provided in the appendix.

\paragraph{Benchmarks.}

To comprehensively evaluate the capabilities of our model, we conducted tests on two distinct categories of benchmarks: long video understanding and long video reasoning.
For long video understanding, we utilized four benchmarks: MLVU~\citep{mlvu}, LongVideoBench~\citep{longvideobench}, VideoMME~\citep{videomme}, and LVBench~\citep{wang2024lvbench}. These benchmarks encompass a variety of tasks designed to assess the model's general video comprehension abilities.
For long video reasoning, we employed three benchmarks that require more than superficial visual analysis: VideoMMLU~\citep{videommlu}, VideoMMMU~\citep{videommmu}, and LongVideoReason-eval~\citep{longvilar1}. These chanllenging benchmarks are specifically designed to evaluate the model's integrated perception and reasoning capabilities.
\subsection{Main Result}
\paragraph{Baselines.}
We compare \ours{} against a wide range of video understanding models, including (1) Proprietary models:   GPT-4o~\citep{gpt4o} and Gemini-1.5-Pro~\citep{gemini}; (2) Open-source VLMs: Video-LLaVA~\citep{videollava}, LLaVA-NeXT-Video~\citep{llavanextvideo}, Video-XL~\citep{videoxl}, VILA-1.5~\citep{lin2024vila}, Kangaroo~\citep{kangaroo}, LongVU~\citep{shen2024longvu}, LongVA~\citep{longva}, LongVILA~\citep{longvila}, LongVILA-R1~\citep{longvilar1}, Video-R1~\citep{videor1} and Qwen2.5-VL~\citep{qwen2_5vl}.

\begin{table}[t]
\centering
 \caption{\textbf{Results on long video benchmarks}.$^\dag$ denotes evaluation results using our own evaluation protocol under max frames of 128. For a fair comparison, our model is evaluated with a maximum of 64 frames in the first round, followed by up to 4 turns requesting a maximum of 16 frames per turn, yielding a total of max 128 frames.}

\begin{adjustbox}{width=\linewidth,center}
\setlength{\tabcolsep}{1.5mm}
\renewcommand{\arraystretch}{1.5}
\begin{tabular}{lrccccccccc}
\toprule  
\multirow{3}{*}{\textbf{Model}} & \multicolumn{1}{c}{\multirow{3}{*}{\centering \textbf{Size}}} & \multicolumn{6}{c}{\textbf{Long Video Understanding}} & \multicolumn{3}{c}{\textbf{Long Video Reasoning}}  \\
\cmidrule(lr){3-8} \cmidrule(lr){9-11}
& & \multicolumn{2}{c}{\textbf{MLVU}} & {\textbf{LongVideoBench } }  &\multicolumn{2}{c}{\textbf{VideoMME}} & {\textbf{LVBench}} & {\textbf{VideoMMLU}} & {\textbf{VideoMMMU}}  &{\textbf{LongVideoReason}}\\ 
& & \textbf{\textit{dev}} &  \textbf{\textit{test}}&\textbf{\textit{val}}&\textbf{\textit{overall}}&\textbf{\textit{long}}&  & \textbf{\textit{quiz} } & &\textbf{\textit{eval} }\\

\midrule
\textit{Proprietary Models} \\
GPT-4o & - & 64.6 & 54.9 & 66.7  & 71.9 & 65.3 & 48.9 &44.9&61.2& 60.7\\
Gemini-1.5-Pro & -  & - &- & 64.0  & 75.0 & 67.4 & 33.1 & -&53.9& 67.3 \\
\midrule
\textit{Open-Source VLMs} \\
Video-LLaVA            &7B &36.2   &30.7   &37.6   &39.9    &    - &    - &    - &    - &    -                                \\
LLaVA-OneVision        &7B &64.7   &47.2   &56.4   &58.3   &46.7          &  -      &33.4   &33.9 &    - \\
LLaVA-NeXT-Video       &7B &    -   &  -     &49.1   &   -                                     &    -   & -      &27.6 &    - &    - \\
Video-XL               &7B &64.9   &45.5   &50.7   &55.5     &    - &    - &    - &    - &    -                               \\
VILA-1.5               &7B &56.7   &  -     &   -    &             -                           &   -    &       -&20.5   &20.9  &    -\\
Kangaroo               &8B &61.0   &-       &54.8   &56.0            &  -     &\uline{39.4}   &    - &    - &    -        \\
LongVU                 &7B &\uline{65.4} &    -   &      - &60.6    &    - &    - &    - &    - &    -                                \\
LongVA                 &7B &56.3   &41.1   &     -  &52.6            &   -    &   -    &  -     &24.0  &    -\\
LongVILA               &7B &   -    &    -   &57.1   &60.1   &    - &    - &    - &    - &    -                                 \\
LongVILA-R1            &7B &      - &  -     &\uline{57.6} &62.4 &53.3 &     -  & -      &   -    &67.9 \\
Video-R1$^\dag$        &7B &65.0   &\uline{49.2} &52.0   &61.1   &51.4   &38.7 &\uline{61.3} &\uline{49.8} &\uline{72.8} \\
Qwen2.5-VL$^\dag$      &7B &58.3   &45.5   &51.0   &\uline{63.5}   &\uline{53.9}   &36.9   &61.0   &48.1   &70.8 \\
\rowcolor{gray!20}\ours$^\dag$&7B &\textbf{68.8} &\textbf{55.8} &\textbf{57.7} &\textbf{65.2} &\textbf{55.8} &\textbf{41.5} &\textbf{67.9} &\textbf{52.2} &\textbf{80.3} \\
\rowcolor{gray!20} $\mathit{\Delta}$ \textit{over base model} 
                         &   &\textcolor{green!60!black}{+10.5} &\textcolor{green!60!black}{+10.3} &\textcolor{green!60!black}{+6.7} &\textcolor{green!60!black}{+1.7} &\textcolor{green!60!black}{+1.9} &\textcolor{green!60!black}{+4.6} &\textcolor{green!60!black}{+6.9} &\textcolor{green!60!black}{+4.1} &\textcolor{green!60!black}{+9.5} \\
\midrule
\bottomrule
\end{tabular}
\end{adjustbox}

\label{tab:main}
\vspace{-1em}
\end{table}

\begin{table}[t]
\centering
\caption{\textbf{Detailed result on MLVU. }ER: Ego Reasoning. NQA: Needle QA, PQA: Plot QA, SQA: Sport QA, AO: Action Order, AC: Action Count, TQA: Tutorial QA, AR: Anomaly Recognition, TR: Topic Reasoning.}

\begin{adjustbox}{width=0.9\linewidth,center}

\label{tab:mlvu}
\begin{tabular}{ll *{10}{c}} 
\toprule
\multirow{2}{*}{Split} & \multirow{2}{*}{Model} & \multicolumn{4}{c}{Single Detail} & \multicolumn{3}{c}{Multi-detail} & \multicolumn{2}{c}{Holistic}  \\
\cmidrule(lr){3-6} \cmidrule(lr){7-9} \cmidrule(lr){10-11} 
& & {ER} & {NQA} & {PQA} & {SQA} & {AO} & {AC} & {TQA} & {AR} & {TR} & {Avg.} \\
\midrule
\multirow{2}{*}{Dev} & Qwen2.5-VL & 47.7 & 65.1 & 65.9 & - & 50.2 & 13.6 & - & \textbf{65.5} & 
\textbf{85.6} & 58.3 \\
& \ours & \textbf{66.8} & \textbf{80.3} & \textbf{72.9} &  - & \textbf{59.8} & \textbf{50.5} & -  & 52.5 & 83.3 & \textbf{68.8} \\
\midrule 
\multirow{2}{*}{Test} & Qwen2.5-VL & 32.1 & 53.3 & 54.0 & \textbf{44.4} & 32.9 & 15.0 & 37.2 & 38.5 & 80.2 & 45.5 \\
& \ours & \textbf{58.5} & \textbf{63.3} & \textbf{64.0} & \textbf{44.4} & \textbf{42.9} & \textbf{28.3} & \textbf{39.5} & \textbf{46.2} & \textbf{89.0} & \textbf{55.8} \\
\bottomrule
\end{tabular}
\end{adjustbox}
\vspace{-1em}
\end{table}

\paragraph{Long Video Understanding.}
Our model demonstrates marked improvements across a range of long video understanding benchmarks, as shown in Table \ref{tab:main}. On MLVU, it achieves scores of 66.8 (dev) and 55.8 (test), yielding substantial gains of +10.5 and +10.3 points over its base model, Qwen2.5-VL. This performance is further validated on LongVideoBench and LVBench, where our model scores 57.7 and 41.5, respectively, outperforming all listed open-source baselines. These results collectively underscore the effectiveness of our adaptive temporal zoom mechanism. Notably, even on benchmarks not exclusively focused on extremely long durations like VideoMME, our method provides a clear performance boost (65.2 overall, 55.8 on long-set) over an already strong baseline. This demonstrates that the learned policy to dynamically \enquote{zoom} in relevant segments is beneficial across various video length.

We present a detailed analysis of our model's performance on the MLVU benchmark in Table \ref{tab:mlvu}. The results clearly show that our method's improvements are most significant on tasks requiring detailed perception. For instance, in the \enquote{Single Detail} category of the dev set, our model shows massive gains in Ego Reasoning (ER, +19.1), Needle QA (NQA, +15.2), and Plot QA (PQA, +7.0). The most significant improvement is seen in the \enquote{Multi-detail} task of Action Count (AC), where our model's score increases from 13.6 to 50.5. This task, which requires counting specific, often rapid actions, directly benefits from the ability to re-sample critical moments at a higher frame rate. Similar substantial gains are observed on the test set in ER (+26.4), NQA (+10.0), and AC (+13.3).

\paragraph{Long Video Reasoning.}
On VideoMMLU and VideoMMMU, our model scores 67.9 and 52.2 respectively, achieving the highest among all open-source models. 
On LongVideoReason-eval, our model achieves a highest score of 80.3, surpassing the performance of powerful proprietary models like GPT-4o (60.7) and Gemini-1.5-Pro (67.3). Notably, our model also outperforms LongVILA-R1, which is trained on the same dataset but with a larger frame budget, highlighting the superior efficiency of our agentic strategy. 
This indicates that the iterative, evidence-gathering process enabled by our agentic strategy allows the model to construct more robust and accurate reasoning chain, which is crucial for tackling complex, knowledge intensive video reasoning tasks.
\begin{figure}[t]
\begin{center}
\begin{subfigure}{0.48\textwidth}
    \includegraphics[scale=0.35]{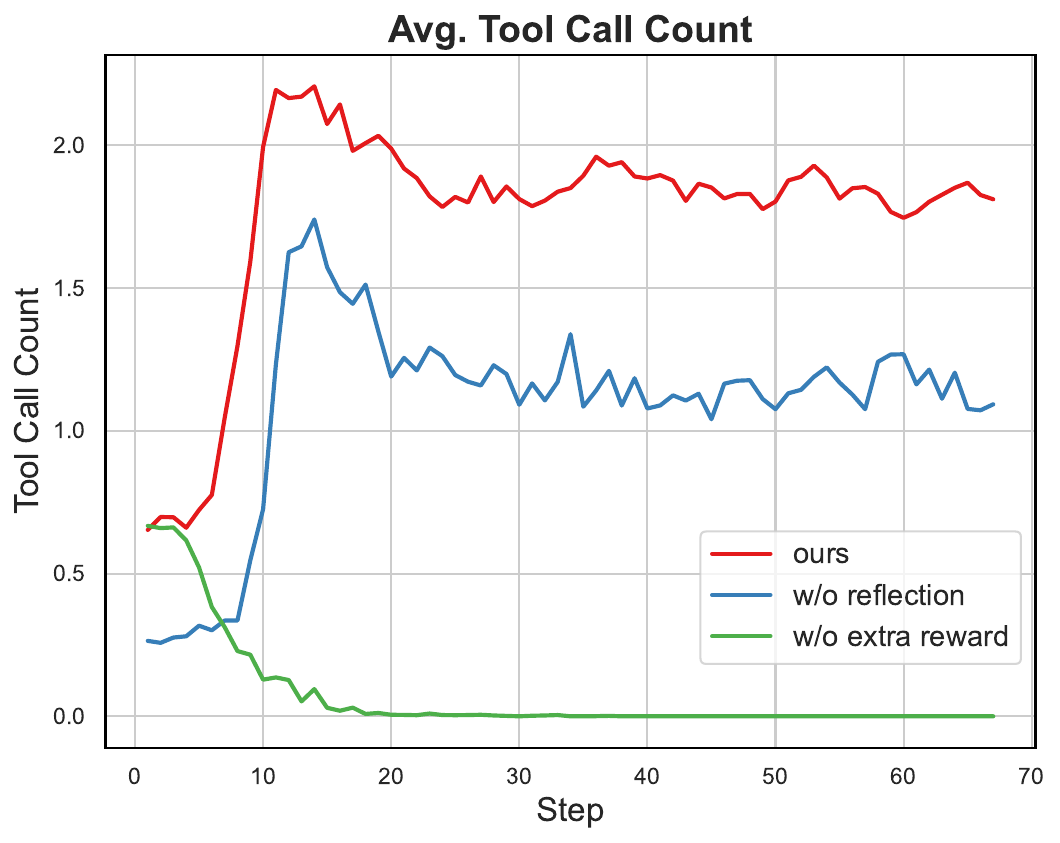}
\end{subfigure}
\begin{subfigure}{0.48\textwidth}
    \includegraphics[scale=0.35]{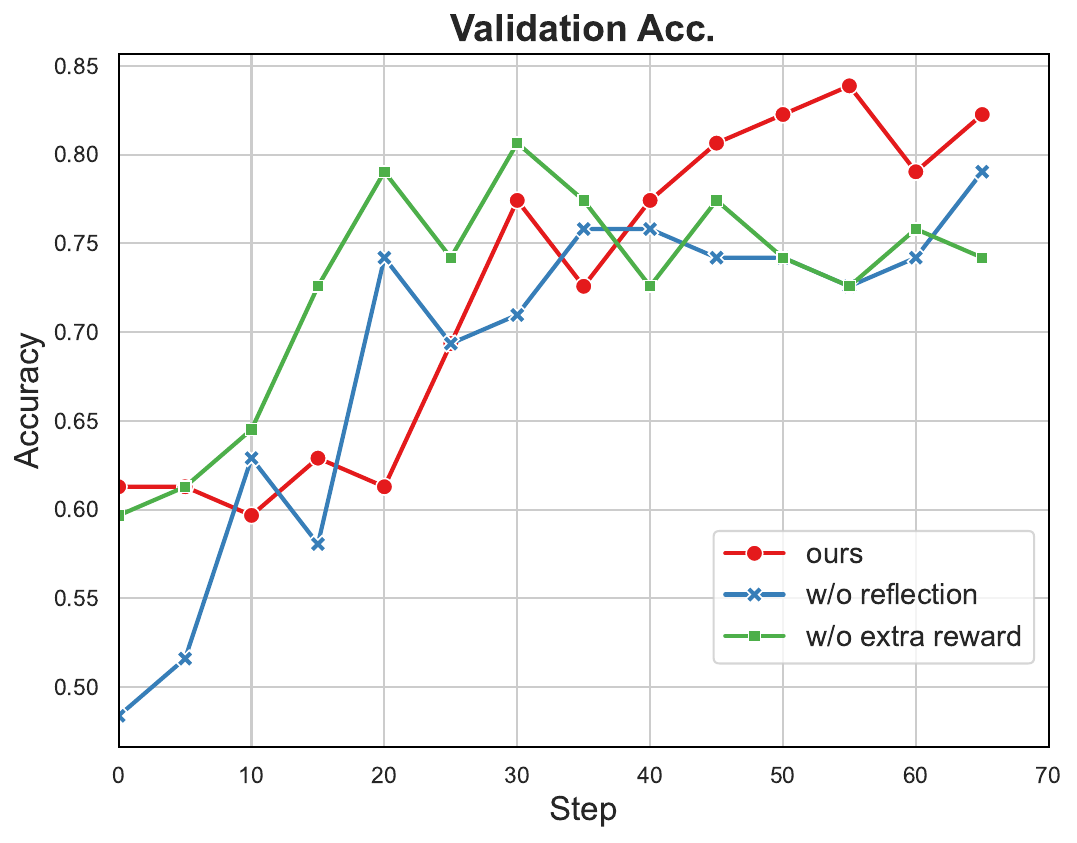}
\end{subfigure}
\caption{\textbf{Training dynamics of ablation baselines.} The left panel shows the average number of tool calls per sample during training. The right panel displays the model performance (e.g., accuracy) on the validation set over the course of training.}
\label{fig:ablation1}
\end{center}
\end{figure}
\begin{table}[t]
\centering
\caption{\textbf{Evaluation of Ablation Baselines}.}
\begin{adjustbox}{width=\linewidth,center}
\setlength{\tabcolsep}{1.5mm}
\renewcommand{\arraystretch}{1.5}
\begin{tabular}{lccccccccc}
\toprule  
\multirow{3}{*}{\textbf{Model}} & \multicolumn{6}{c}{\textbf{Long Video Understanding}} & \multicolumn{3}{c}{\textbf{Long Video Reasoning}}  \\
\cmidrule(lr){2-7} \cmidrule(lr){8-10}
&  \multicolumn{2}{c}{\textbf{MLVU}} & {\textbf{LongVideoBench } }  &\multicolumn{2}{c}{\textbf{VideoMME}} & {\textbf{LVBench}} & {\textbf{VideoMMLU}} & {\textbf{VideoMMMU}}  &{\textbf{LongVideoReason}}\\ 
&  \textbf{\textit{dev}} &  \textbf{\textit{test}}&\textbf{\textit{val}}&\textbf{\textit{overall}}&\textbf{\textit{long}}&  & \textbf{\textit{quiz} } & &\textbf{\textit{eval} }\\

\midrule
\ours  &\textbf{68.8}   &\textbf{55.8}   &\textbf{57.7}   &\textbf{65.2}   &\textbf{55.8}   &\textbf{41.5}   &67.9   &52.2   &\textbf{80.3}\\
\midrule
w/o RL  &56.4   &45.6   &42.0   &54.4   &44.2   &26.0   &63.5   &46.6   &63.3\\ 
w/o $R_{tool}$  &67.5   &52.2   &56.2   &62.5   &52.5   &40.6   &63.6   & \textbf{53.8}  &79.9\\
w/o cold-start &57.0    &42.8  &43.5   &53.5   &46.6   & 35.5  &63.9  & 43.6    &59.6\\
w/o reflection  &\textcolor{blue}{67.0}   &\textcolor{blue}{53.2}   &\textcolor{blue}{54.8}  &\textcolor{blue}{58.7}     &\textcolor{blue}{47.4}   &\textcolor{blue}{40.9}  &\textcolor{blue}{70.1} &\textcolor{blue}{52.2}   &\textcolor{blue}{75.1}\\

\bottomrule
\end{tabular}
\end{adjustbox}

\label{tab:ablation}
\end{table}

\subsection{Ablation Study}
\paragraph{Effectiveness of Key Components.}
To validate the contribution of each key component in our framework, we conduct a comprehensive ablation study, with results summarized in Table \ref{tab:ablation} and training dynamics shown in Figure \ref{fig:ablation1}. \textcolor{blue}{For a fair comparison, all ablated models except \enquote{w/o cold-start} were trained using the same amount of SFT data as our final model.}
The w/o RL model, trained only via supervised fine-tuning, suffers a catastrophic performance drop across all benchmarks (e.g., -17.0 on LongVideoReason), confirming that RL is essential for learning an effective tool-use policy. Similarly, the w/o cold-start model, which skips our curated SFT stage, fails to converge to a meaningful policy, highlighting the necessity of a strong initialization. Within the cold-start process, removing reflection data (w/o reflection) causes the model to adopt a shallow, simple strategy, where the average tool call count stabilizes at about 1.0, limiting its ability to tackle complex problems. In contrast, our full method learns to make nearly two calls on average, enabling deeper investigation and achieving higher accuracy in the validation set. Finally, removing the conditional tool-use bonus (w/o $R_{tool}$) leads to \enquote{policy collapse}, where the agent's tool usage trends towards zero during training, as it fails to discover the tool's utility without explicit encouragement. Each ablation results in significantly lower performance on various benchmarks, demonstrating that all components are necessary to achieve the final performance.

\begin{figure}[t]
\begin{center}
\begin{subfigure}{0.32\textwidth}
    \includegraphics[width=\textwidth]{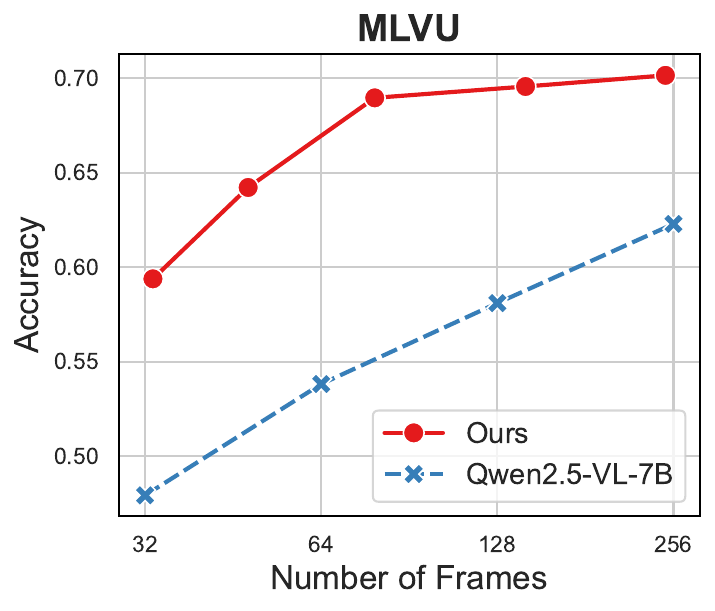}
\end{subfigure}
\begin{subfigure}{0.32\textwidth}
    \includegraphics[width=\textwidth]{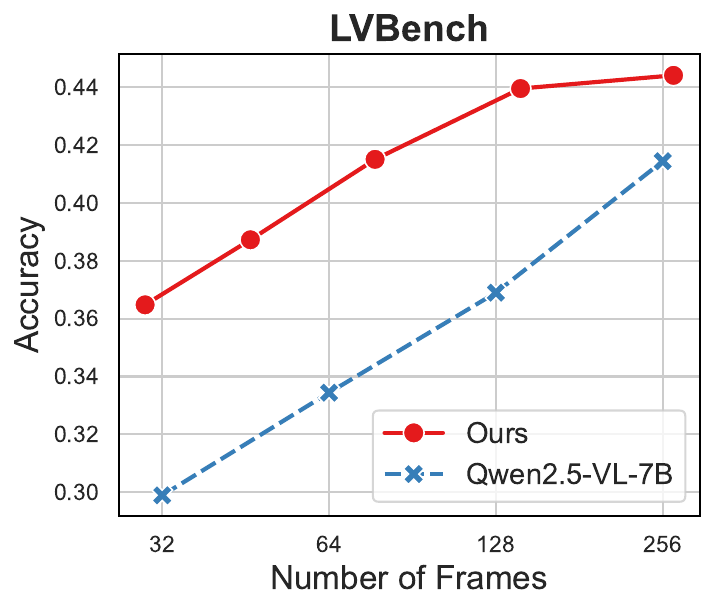}
\end{subfigure}
\begin{subfigure}{0.32\textwidth}
    \includegraphics[width=\textwidth]{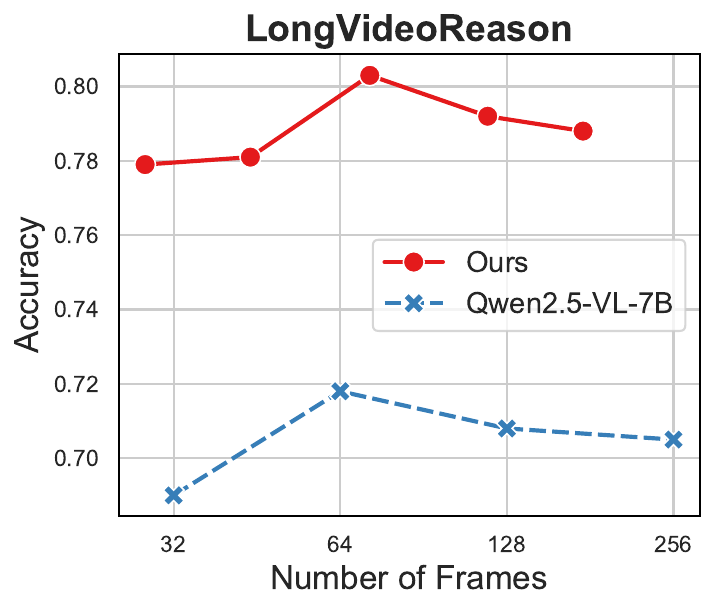}
\end{subfigure}
\caption{\textbf{Performance comparison with varying frame budgets.} We compare our model against the Qwen2.5-VL-7B baseline.  The x-axis(log scale) represents the fixed frame budget for the baseline and the average number of frames actually used by \ours~on each benchmark. }
\label{fig:ablation2}
\end{center}
\vspace{-1em}
\end{figure}

\paragraph{Performance Across Various Frame Budgets.}
To further investigate the efficiency of \ours, we analyze the performance of our model and the base model under various frame budgets. As illustrated in Figure \ref{fig:ablation2}, we plot the accuracy of the model against the number of frames processed. For the baseline model, this x-axis represents a fixed, uniformly sampled frame budget. For our model, it represents the actual average number of frames consumed per dataset, a result of its dynamic decision-making process. 
The results clearly demonstrate our model's superior efficiency. On MLVU, our model achieves 0.64 accuracy using only 48 frames on average, surpassing the baseline's 0.581 accuracy at a much larger 128 frame budget. This trend holds on LVBench, where our model using 77 frames outperforms the baseline using 256 frames. 
Furthermore, on the LongVideoReason benchmark, our model and the baseline model both peaks at around 64 frames, suggesting that complex reasoning tasks may not benefit from increasing visual information, which can introduce noise. However, within this optimal frame window, our model's peak accuracy of 0.803 significantly surpasses the base model's peak of 0.718. This performance gap underscores our model's stronger reasoning capability enabled by its agentic policy.

\begin{wraptable}{r}{0.5\textwidth}
\vspace{-1.5em}
\caption{\textbf{Performance comparison when combining with an external frame selector.} Results are evaluated using our protocol under a consistent setting.}
\label{tab:frmselector}
\begin{adjustbox}{width=0.45\textwidth,center}
    \centering
    \begin{tabular}{rcc}
        \toprule
        Model & MLVU & LongVideoBench \\
        \midrule
        Qwen2.5VL & 58.1 & 51.0 \\
        \quad \textit{+tspo} & 68.1 & 54.9 \\
        \ours & 68.8 & 57.7 \\
        \quad \textit{+tspo} & \textbf{70.8} & \textbf{60.7} \\
        \bottomrule
    \end{tabular}
\end{adjustbox}
\end{wraptable}
\vspace{-1em}
\paragraph{Combining with a Frame Selector.}
Our primary method uses uniform sampling for the initial overview to ensure a global and unbiased starting point, we also investigate whether our agentic framework can be combined with more sophisticated frame selectors. To test this, we replace the initial uniformly sampled frames with the output of the output of TSPO-0.4B~\citep{tang2025tspo}. The results presented in Table \ref{tab:frmselector} shows that providing a more intelligently selected initial overview further boosts our model's performance by +2.0 on MLVU and +3.0 on LongVideoBench. This demonstrates the flexibility and transferability of our approach; the learned policy effectively leverages the improved starting point to conduct an even more efficient and accurate investigation of the video.

\section{Conclusion}
In this work, we propose \ours~to address the critical challenge of long video understanding in MLLMs. We empower the MLLM to become an active agent capable of utilizing external tool to investigate long videos more effectively and efficiently through a carefully designed two-stage training process. 

Our experimental results robustly validated our approach. The ablation studies confirmed that each component—the cold-start initialization, the reflection data, the RL optimization, and the conditional reward bonus—was indispensable for achieving final performance. Our model not only achieves strong performance across numerous long video benchmarks, but also demonstrated superior frame efficiency, outperforming stronger baselines while using significantly fewer frames. This demonstrates the effectiveness of our agentic strategy in enhancing the perception and reasoning capabilities of MLLMs for long videos.

\bibliography{iclr2026_conference}

@misc{gpt4o,
  author = {OpenAI},
  title = {GPT-4o},
  howpublished = {\url{https://openai.com/index/hello-gpt-4o/}},
  month = {May},
  year = {2024}
}

@article{gemini,
  author       = {Machel Reid and
                  Nikolay Savinov and
                  Denis Teplyashin and
                  Dmitry Lepikhin and
                  Timothy P. Lillicrap and
                  Jean{-}Baptiste Alayrac and
                  Radu Soricut and
                  Angeliki Lazaridou and
                  Orhan Firat and
                  Julian Schrittwieser and
                  Ioannis Antonoglou and
                  Rohan Anil and
                  Sebastian Borgeaud and
                  Andrew M. Dai and
                  Katie Millican and
                  Ethan Dyer and
                  Mia Glaese and
                  Thibault Sottiaux and
                  Benjamin Lee and
                  Fabio Viola and
                  Malcolm Reynolds and
                  Yuanzhong Xu and
                  James Molloy and
                  Jilin Chen and
                  Michael Isard and
                  Paul Barham and
                  Tom Hennigan and
                  Ross McIlroy and
                  Melvin Johnson and
                  Johan Schalkwyk and
                  Eli Collins and
                  Eliza Rutherford and
                  Erica Moreira and
                  Kareem Ayoub and
                  Megha Goel and
                  Clemens Meyer and
                  Gregory Thornton and
                  Zhen Yang and
                  Henryk Michalewski and
                  Zaheer Abbas and
                  Nathan Schucher and
                  Ankesh Anand and
                  Richard Ives and
                  James Keeling and
                  Karel Lenc and
                  Salem Haykal and
                  Siamak Shakeri and
                  Pranav Shyam and
                  Aakanksha Chowdhery and
                  Roman Ring and
                  Stephen Spencer and
                  Eren Sezener and
                  et al.},
  title        = {Gemini 1.5: Unlocking multimodal understanding across millions of
                  tokens of context},
  journal      = {CoRR},
  volume       = {abs/2403.05530},
  year         = {2024}
}

@article{comanici2025gemini2.5,
  title={Gemini 2.5: Pushing the frontier with advanced reasoning, multimodality, long context, and next generation agentic capabilities},
  author={Comanici, Gheorghe and Bieber, Eric and Schaekermann, Mike and Pasupat, Ice and Sachdeva, Noveen and Dhillon, Inderjit and Blistein, Marcel and Ram, Ori and Zhang, Dan and Rosen, Evan and others},
  journal={arXiv preprint arXiv:2507.06261},
  year={2025}
}

@article{videollava,
  title={Video-llava: Learning united visual representation by alignment before projection},
  author={Lin, Bin and Ye, Yang and Zhu, Bin and Cui, Jiaxi and Ning, Munan and Jin, Peng and Yuan, Li},
  journal={arXiv preprint arXiv:2311.10122},
  year={2023}
}

@article{longvideobench,
  title={Longvideobench: A benchmark for long-context interleaved video-language understanding},
  author={Wu, Haoning and Li, Dongxu and Chen, Bei and Li, Junnan},
  journal={arXiv preprint arXiv:2407.15754},
  year={2024}
}

@article{mlvu,
  title={MLVU: A Comprehensive Benchmark for Multi-Task Long Video Understanding},
  author={Zhou, Junjie and Shu, Yan and Zhao, Bo and Wu, Boya and Xiao, Shitao and Yang, Xi and Xiong, Yongping and Zhang, Bo and Huang, Tiejun and Liu, Zheng},
  journal={arXiv preprint arXiv:2406.04264},
  year={2024}
}

@article{videomme,
  title={Video-MME: The First-Ever Comprehensive Evaluation Benchmark of Multi-modal LLMs in Video Analysis},
  author={Fu, Chaoyou and Dai, Yuhan and Luo, Yondong and Li, Lei and Ren, Shuhuai and Zhang, Renrui and Wang, Zihan and Zhou, Chenyu and Shen, Yunhang and Zhang, Mengdan and others},
  journal={arXiv preprint arXiv:2405.21075},
  year={2024}
}

@misc{llavanextvideo,
  title={LLaVA-NeXT: A Strong Zero-shot Video Understanding Model},
  url={https://llava-vl.github.io/blog/2024-04-30-llava-next-video/},
  author={Zhang, Yuanhan and Li, Bo and Liu, haotian and Lee, Yong jae and Gui, Liangke and Fu, Di and Feng, Jiashi and Liu, Ziwei and Li, Chunyuan},
  month={April},
  year={2024}
}

@article{longvila,
  title={Longvila: Scaling long-context visual language models for long videos},
  author={Xue, Fuzhao and Chen, Yukang and Li, Dacheng and Hu, Qinghao and Zhu, Ligeng and Li, Xiuyu and Fang, Yunhao and Tang, Haotian and Yang, Shang and Liu, Zhijian and others},
  journal={arXiv preprint arXiv:2408.10188},
  year={2024}
}

@article{longva,
  author       = {Peiyuan Zhang and
                  Kaichen Zhang and
                  Bo Li and
                  Guangtao Zeng and
                  Jingkang Yang and
                  Yuanhan Zhang and
                  Ziyue Wang and
                  Haoran Tan and
                  Chunyuan Li and
                  Ziwei Liu},
  title        = {Long Context Transfer from Language to Vision},
  journal      = {CoRR},
  volume       = {abs/2406.16852},
  year         = {2024}
}

@article{videoxl,
  title={Video-XL: Extra-Long Vision Language Model for Hour-Scale Video Understanding},
  author={Shu, Yan and Zhang, Peitian and Liu, Zheng and Qin, Minghao and Zhou, Junjie and Huang, Tiejun and Zhao, Bo},
  journal={arXiv preprint arXiv:2409.14485},
  year={2024}
}

@article{kangaroo,
  title={Kangaroo: A powerful video-language model supporting long-context video input},
  author={Liu, Jiajun and Wang, Yibing and Ma, Hanghang and Wu, Xiaoping and Ma, Xiaoqi and Wei, Xiaoming and Jiao, Jianbin and Wu, Enhua and Hu, Jie},
  journal={arXiv preprint arXiv:2408.15542},
  year={2024}
}

@article{qwen2_5vl,
  title={Qwen2. 5-vl technical report},
  author={Bai, Shuai and Chen, Keqin and Liu, Xuejing and Wang, Jialin and Ge, Wenbin and Song, Sibo and Dang, Kai and Wang, Peng and Wang, Shijie and Tang, Jun and others},
  journal={arXiv preprint arXiv:2502.13923},
  year={2025}
}

@inproceedings{lin2024vila,
  title={Vila: On pre-training for visual language models},
  author={Lin, Ji and Yin, Hongxu and Ping, Wei and Molchanov, Pavlo and Shoeybi, Mohammad and Han, Song},
  booktitle={Proceedings of the IEEE/CVF conference on computer vision and pattern recognition},
  pages={26689--26699},
  year={2024}
}

@article{shen2024longvu,
  title={Longvu: Spatiotemporal adaptive compression for long video-language understanding},
  author={Shen, Xiaoqian and Xiong, Yunyang and Zhao, Changsheng and Wu, Lemeng and Chen, Jun and Zhu, Chenchen and Liu, Zechun and Xiao, Fanyi and Varadarajan, Balakrishnan and Bordes, Florian and others},
  journal={arXiv preprint arXiv:2410.17434},
  year={2024}
}

@article{longvilar1,
  title={Scaling rl to long videos},
  author={Chen, Yukang and Huang, Wei and Shi, Baifeng and Hu, Qinghao and Ye, Hanrong and Zhu, Ligeng and Liu, Zhijian and Molchanov, Pavlo and Kautz, Jan and Qi, Xiaojuan and others},
  journal={arXiv preprint arXiv:2507.07966},
  year={2025}
}

@article{videor1,
  title={Video-r1: Reinforcing video reasoning in mllms},
  author={Feng, Kaituo and Gong, Kaixiong and Li, Bohao and Guo, Zonghao and Wang, Yibing and Peng, Tianshuo and Wu, Junfei and Zhang, Xiaoying and Wang, Benyou and Yue, Xiangyu},
  journal={arXiv preprint arXiv:2503.21776},
  year={2025}
}

@article{wang2024lvbench,
  title={Lvbench: An extreme long video understanding benchmark},
  author={Wang, Weihan and He, Zehai and Hong, Wenyi and Cheng, Yean and Zhang, Xiaohan and Qi, Ji and Gu, Xiaotao and Huang, Shiyu and Xu, Bin and Dong, Yuxiao and others},
  journal={arXiv preprint arXiv:2406.08035},
  year={2024}
}

@article{videommlu,
  title={Video-mmlu: A massive multi-discipline lecture understanding benchmark},
  author={Song, Enxin and Chai, Wenhao and Xu, Weili and Xie, Jianwen and Liu, Yuxuan and Wang, Gaoang},
  journal={arXiv preprint arXiv:2504.14693},
  year={2025}
}

@article{videommmu,
  title={Video-mmmu: Evaluating knowledge acquisition from multi-discipline professional videos},
  author={Hu, Kairui and Wu, Penghao and Pu, Fanyi and Xiao, Wang and Zhang, Yuanhan and Yue, Xiang and Li, Bo and Liu, Ziwei},
  journal={arXiv preprint arXiv:2501.13826},
  year={2025}
}

@article{verl,
  title   = {HybridFlow: A Flexible and Efficient RLHF Framework},
  author  = {Guangming Sheng and Chi Zhang and Zilingfeng Ye and Xibin Wu and Wang Zhang and Ru Zhang and Yanghua Peng and Haibin Lin and Chuan Wu},
  year    = {2024},
  journal = {arXiv preprint arXiv: 2409.19256}
}

@inproceedings{zheng2024llamafactory,
  title={LlamaFactory: Unified Efficient Fine-Tuning of 100+ Language Models},
  author={Yaowei Zheng and Richong Zhang and Junhao Zhang and Yanhan Ye and Zheyan Luo and Zhangchi Feng and Yongqiang Ma},
  booktitle={Proceedings of the 62nd Annual Meeting of the Association for Computational Linguistics (Volume 3: System Demonstrations)},
  address={Bangkok, Thailand},
  publisher={Association for Computational Linguistics},
  year={2024},
  url={http://arxiv.org/abs/2403.13372}
}

@article{yu2025dapo,
  title={Dapo: An open-source llm reinforcement learning system at scale},
  author={Yu, Qiying and Zhang, Zheng and Zhu, Ruofei and Yuan, Yufeng and Zuo, Xiaochen and Yue, Yu and Dai, Weinan and Fan, Tiantian and Liu, Gaohong and Liu, Lingjun and others},
  journal={arXiv preprint arXiv:2503.14476},
  year={2025}
}

@article{shao2024deepseekmath,
  title={Deepseekmath: Pushing the limits of mathematical reasoning in open language models},
  author={Shao, Zhihong and Wang, Peiyi and Zhu, Qihao and Xu, Runxin and Song, Junxiao and Bi, Xiao and Zhang, Haowei and Zhang, Mingchuan and Li, YK and Wu, Yang and others},
  journal={arXiv preprint arXiv:2402.03300},
  year={2024}
}

@article{llavavideo,
  title={Video instruction tuning with synthetic data},
  author={Zhang, Yuanhan and Wu, Jinming and Li, Wei and Li, Bo and Ma, Zejun and Liu, Ziwei and Li, Chunyuan},
  journal={arXiv preprint arXiv:2410.02713},
  year={2024}
}

@inproceedings{hu2025m,
  title={M-LLM based video frame selection for efficient video understanding},
  author={Hu, Kai and Gao, Feng and Nie, Xiaohan and Zhou, Peng and Tran, Son and Neiman, Tal and Wang, Lingyun and Shah, Mubarak and Hamid, Raffay and Yin, Bing and others},
  booktitle={Proceedings of the Computer Vision and Pattern Recognition Conference},
  pages={13702--13712},
  year={2025}
}

@article{tang2025tspo,
  title={TSPO: Temporal Sampling Policy Optimization for Long-form Video Language Understanding},
  author={Tang, Canhui and Han, Zifan and Sun, Hongbo and Zhou, Sanping and Zhang, Xuchong and Wei, Xin and Yuan, Ye and Xu, Jinglin and Sun, Hao},
  journal={arXiv preprint arXiv:2508.04369},
  year={2025}
}

@article{guo2025deepseek,
  title={Deepseek-r1: Incentivizing reasoning capability in llms via reinforcement learning},
  author={Guo, Daya and Yang, Dejian and Zhang, Haowei and Song, Junxiao and Zhang, Ruoyu and Xu, Runxin and Zhu, Qihao and Ma, Shirong and Wang, Peiyi and Bi, Xiao and others},
  journal={arXiv preprint arXiv:2501.12948},
  year={2025}
}

@article{meng2025mm,
  title={Mm-eureka: Exploring the frontiers of multimodal reasoning with rule-based reinforcement learning},
  author={Meng, Fanqing and Du, Lingxiao and Liu, Zongkai and Zhou, Zhixiang and Lu, Quanfeng and Fu, Daocheng and Han, Tiancheng and Shi, Botian and Wang, Wenhai and He, Junjun and others},
  journal={arXiv preprint arXiv:2503.07365},
  year={2025}
}

@article{zheng2025deepeyes,
  title={DeepEyes: Incentivizing" Thinking with Images" via Reinforcement Learning},
  author={Zheng, Ziwei and Yang, Michael and Hong, Jack and Zhao, Chenxiao and Xu, Guohai and Yang, Le and Shen, Chao and Yu, Xing},
  journal={arXiv preprint arXiv:2505.14362},
  year={2025}
}

@article{su2025pixel,
  title={Pixel reasoner: Incentivizing pixel-space reasoning with curiosity-driven reinforcement learning},
  author={Su, Alex and Wang, Haozhe and Ren, Weiming and Lin, Fangzhen and Chen, Wenhu},
  journal={arXiv preprint arXiv:2505.15966},
  year={2025}
}

@article{wu2025mmsearch,
  title={MMSearch-R1: Incentivizing LMMs to Search},
  author={Wu, Jinming and Deng, Zihao and Li, Wei and Liu, Yiding and You, Bo and Li, Bo and Ma, Zejun and Liu, Ziwei},
  journal={arXiv preprint arXiv:2506.20670},
  year={2025}
}

@article{liu2025seg,
  title={Seg-zero: Reasoning-chain guided segmentation via cognitive reinforcement},
  author={Liu, Yuqi and Peng, Bohao and Zhong, Zhisheng and Yue, Zihao and Lu, Fanbin and Yu, Bei and Jia, Jiaya},
  journal={arXiv preprint arXiv:2503.06520},
  year={2025}
}

@article{liu2025video,
  title={Video-xl-pro: Reconstructive token compression for extremely long video understanding},
  author={Liu, Xiangrui and Shu, Yan and Liu, Zheng and Li, Ao and Tian, Yang and Zhao, Bo},
  journal={arXiv preprint arXiv:2503.18478},
  year={2025}
}

@article{yan2025crosslmm,
  title={Crosslmm: Decoupling long video sequences from lmms via dual cross-attention mechanisms},
  author={Yan, Shilin and Han, Jiaming and Tsai, Joey and Xue, Hongwei and Fang, Rongyao and Hong, Lingyi and Guo, Ziyu and Zhang, Ray},
  journal={arXiv preprint arXiv:2505.17020},
  year={2025}
}

@article{yuan2025videodeepresearch,
  title={VideoDeepResearch: Long Video Understanding With Agentic Tool Using},
  author={Yuan, Huaying and Liu, Zheng and Zhou, Junjie and Qian, Hongjin and Wen, Ji-Rong and Dou, Zhicheng},
  journal={arXiv preprint arXiv:2506.10821},
  year={2025}
}

@article{zhang2025deep,
  title={Deep Video Discovery: Agentic Search with Tool Use for Long-form Video Understanding},
  author={Zhang, Xiaoyi and Jia, Zhaoyang and Guo, Zongyu and Li, Jiahao and Li, Bin and Li, Houqiang and Lu, Yan},
  journal={arXiv preprint arXiv:2505.18079},
  year={2025}
}

@article{wang2025vl,
  title={Vl-rethinker: Incentivizing self-reflection of vision-language models with reinforcement learning},
  author={Wang, Haozhe and Qu, Chao and Huang, Zuming and Chu, Wei and Lin, Fangzhen and Chen, Wenhu},
  journal={arXiv preprint arXiv:2504.08837},
  year={2025}
}

@article{Kietzmann2018DeepNN,
  title={Deep Neural Networks in Computational Neuroscience},
  author={Tim C Kietzmann and Patrick McClure and Nikolaus Kriegeskorte},
  journal={bioRxiv},
  year={2018},
  url={https://api.semanticscholar.org/CorpusID:195946461}
}

@inproceedings{Zhang2023VideoLLaMAAI,
  title={Video-LLaMA: An Instruction-tuned Audio-Visual Language Model for Video Understanding},
  author={Hang Zhang and Xin Li and Lidong Bing},
  booktitle={Conference on Empirical Methods in Natural Language Processing},
  year={2023},
  url={https://api.semanticscholar.org/CorpusID:259075356}
}

@article{Chen2024ExpandingPB,
  title={Expanding Performance Boundaries of Open-Source Multimodal Models with Model, Data, and Test-Time Scaling},
  author={Zhe Chen and Weiyun Wang and Yue Cao and Yangzhou Liu and Zhangwei Gao and Erfei Cui and Jinguo Zhu and Shenglong Ye and Hao Tian and Zhaoyang Liu and Lixin Gu and Xuehui Wang and Qingyun Li and Yiming Ren and Zixuan Chen and Jiapeng Luo and Jiahao Wang and Tan Jiang and Bo Wang and Conghui He and Botian Shi and Xingcheng Zhang and Han Lv and Yi Wang and Wenqi Shao and Pei Chu and Zhongying Tu and Tong He and Zhiyong Wu and Hui Deng and Jiaye Ge and Kaiming Chen and Min Dou and Lewei Lu and Xizhou Zhu and Tong Lu and Dahu Lin and Yunfeng Qiao and Jifeng Dai and Wenhai Wang},
  journal={ArXiv},
  year={2024},
  volume={abs/2412.05271},
  url={https://api.semanticscholar.org/CorpusID:274581884}
}

@inproceedings{vllm,
  title={Efficient Memory Management for Large Language Model Serving with PagedAttention},
  author={Woosuk Kwon and Zhuohan Li and Siyuan Zhuang and Ying Sheng and Lianmin Zheng and Cody Hao Yu and Joseph E. Gonzalez and Hao Zhang and Ion Stoica},
  booktitle={Proceedings of the ACM SIGOPS 29th Symposium on Operating Systems Principles},
  year={2023}
}

@article{yu2024frame,
  title={Frame-voyager: Learning to query frames for video large language models},
  author={Yu, Sicheng and Jin, Chengkai and Wang, Huanyu and Chen, Zhenghao and Jin, Sheng and Zuo, Zhongrong and Xu, Xiaolei and Sun, Zhenbang and Zhang, Bingni and Wu, Jiawei and others},
  journal={arXiv preprint arXiv:2410.03226},
  year={2024}
}

@inproceedings{wang2024vila,
  title={Vila: Efficient video-language alignment for video question answering},
  author={Wang, Xijun and Liang, Junbang and Wang, Chun-Kai and Deng, Kenan and Lou, Yu and Lin, Ming C and Yang, Shan},
  booktitle={European Conference on Computer Vision},
  pages={186--204},
  year={2024},
  organization={Springer}
}

@article{lsdbench,
  title={Does Your Vision-Language Model Get Lost in the Long Video Sampling Dilemma?},
  author={Qu, Tianyuan and Tang, Longxiang and Peng, Bohao and Yang, Senqiao and Yu, Bei and Jia, Jiaya},
  journal={arXiv preprint arXiv:2503.12496},
  year={2025}
}

@article{team2025kimi,
  title={Kimi k1. 5: Scaling reinforcement learning with llms},
  author={Team, Kimi and Du, Angang and Gao, Bofei and Xing, Bowei and Jiang, Changjiu and Chen, Cheng and Li, Cheng and Xiao, Chenjun and Du, Chenzhuang and Liao, Chonghua and others},
  journal={arXiv preprint arXiv:2501.12599},
  year={2025}
}

@article{tan2025reason,
  title={Reason-rft: Reinforcement fine-tuning for visual reasoning},
  author={Tan, Huajie and Ji, Yuheng and Hao, Xiaoshuai and Lin, Minglan and Wang, Pengwei and Wang, Zhongyuan and Zhang, Shanghang},
  journal={arXiv preprint arXiv:2503.20752},
  year={2025}
}

@article{jaech2024openaio1,
  title={Openai o1 system card},
  author={Jaech, Aaron and Kalai, Adam and Lerer, Adam and Richardson, Adam and El-Kishky, Ahmed and Low, Aiden and Helyar, Alec and Madry, Aleksander and Beutel, Alex and Carney, Alex and others},
  journal={arXiv preprint arXiv:2412.16720},
  year={2024}
}

@article{yang2025qwen3,
  title={Qwen3 technical report},
  author={Yang, An and Li, Anfeng and Yang, Baosong and Zhang, Beichen and Hui, Binyuan and Zheng, Bo and Yu, Bowen and Gao, Chang and Huang, Chengen and Lv, Chenxu and others},
  journal={arXiv preprint arXiv:2505.09388},
  year={2025}
}

@article{cai2024temporalbench,
  title={Temporalbench: Benchmarking fine-grained temporal understanding for multimodal video models},
  author={Cai, Mu and Tan, Reuben and Zhang, Jianrui and Zou, Bocheng and Zhang, Kai and Yao, Feng and Zhu, Fangrui and Gu, Jing and Zhong, Yiwu and Shang, Yuzhang and others},
  journal={arXiv preprint arXiv:2410.10818},
  year={2024}
}

@article{chai2024auroracap,
  title={Auroracap: Efficient, performant video detailed captioning and a new benchmark},
  author={Chai, Wenhao and Song, Enxin and Du, Yilun and Meng, Chenlin and Madhavan, Vashisht and Bar-Tal, Omer and Hwang, Jenq-Neng and Xie, Saining and Manning, Christopher D},
  journal={arXiv preprint arXiv:2410.03051},
  year={2024}
}

@article{yi2019clevrer,
  title={Clevrer: Collision events for video representation and reasoning},
  author={Yi, Kexin and Gan, Chuang and Li, Yunzhu and Kohli, Pushmeet and Wu, Jiajun and Torralba, Antonio and Tenenbaum, Joshua B},
  journal={arXiv preprint arXiv:1910.01442},
  year={2019}
}

@article{liu2024tempcompass,
  title={Tempcompass: Do video llms really understand videos?},
  author={Liu, Yuanxin and Li, Shicheng and Liu, Yi and Wang, Yuxiang and Ren, Shuhuai and Li, Lei and Chen, Sishuo and Sun, Xu and Hou, Lu},
  journal={arXiv preprint arXiv:2403.00476},
  year={2024}
}

@article{ye2025limo,
  title={Limo: Less is more for reasoning},
  author={Ye, Yixin and Huang, Zhen and Xiao, Yang and Chern, Ethan and Xia, Shijie and Liu, Pengfei},
  journal={arXiv preprint arXiv:2502.03387},
  year={2025}
}
\bibliographystyle{iclr2026_conference}

\appendix
\clearpage

\section{More Implement Details}
\label{sec:impl_details}

\paragraph{Evaluation Details.}
We evaluate our model and baselines under a consistent setting with a maximum of 128 frames and a resolution corresponding to 100,352 pixels per frame. For inference, we employed the vLLM framework \citep{vllm} with the temperature parameter set to 0 to ensure deterministic outputs.

For the VideoMMLU benchmark, answers are scored by GPT-4o using the official prompt and the final score is computed as the average score of three disciplines.
\paragraph{Training Details.} 
We show the key training hyperparameters in Table \ref{tab:keyhyperpara}.
\begin{table}[h]
\caption{Key Hyperparameters}
\label{tab:keyhyperpara}
\begin{subtable}[t]{0.48\textwidth}
\caption{SFT stage}
\centering
\begin{tabular}{@{}ll@{}}
\toprule
\textbf{Hyperparameter} & \textbf{Value} \\
\midrule
Train epochs & \texttt{1} \\
Train batch size & \texttt{64} \\
Learning rate & \texttt{5e-5} \\
Learning rate scheluder & \texttt{cosine} \\
Warmup ratio & \texttt{0.1} \\
Freeze vision encoder & \texttt{true} \\
\bottomrule
\end{tabular}
\end{subtable}
\begin{subtable}[t]{0.48\textwidth}
\caption{RL stage}
\centering
\begin{tabular}{@{}ll@{}}
\toprule
\textbf{Hyperparameter} & \textbf{Value} \\
\midrule
Max total response length & \texttt{32768} \\
Rollout temperature & \texttt{1.0} \\
Max interaction turns & \texttt{5} \\
Train batch size & \texttt{128} \\
PPO mini batch size & \texttt{32} \\
Rollouts per prompt ($n$) & \texttt{16} \\
Clip ratio (low / high) & \texttt{0.2 / 0.27} \\
Entropy coefficient & \texttt{0.001} \\
KL coefficient ($\beta$) & \texttt{0.001} \\
Learning rate & \texttt{1e-6} \\
Reward weight (acc/format/tool) & \texttt{0.9/0.1/0.5}\\
\bottomrule
\end{tabular}
\end{subtable}
\end{table}

The SFT training is conducted on 8$\times$H100 GPUs for  $\sim$6h, RL training is conducted on 16$\times$H100 GPUs for  $\sim$45h.

Figure \ref{fig:stats} shows key statistics of our cold-start dataset. The left panel shows the distribution of total token lengths per trajectory, indicating a wide variety of response lengths that cover both simple and complex reasoning chains. The right panel illustrates the distribution of interaction rounds (i.e., the number of tool calls), showing that the dataset contains a significant number of multi-step examples.

\paragraph{Prompt Template.}
We provide the detailed prompt used for training and cold-start data synthesization as follows: 

\begin{figure}[htbp]
    \centering
    \begin{subfigure}[b]{0.48\textwidth}
        \includegraphics[width=\textwidth]{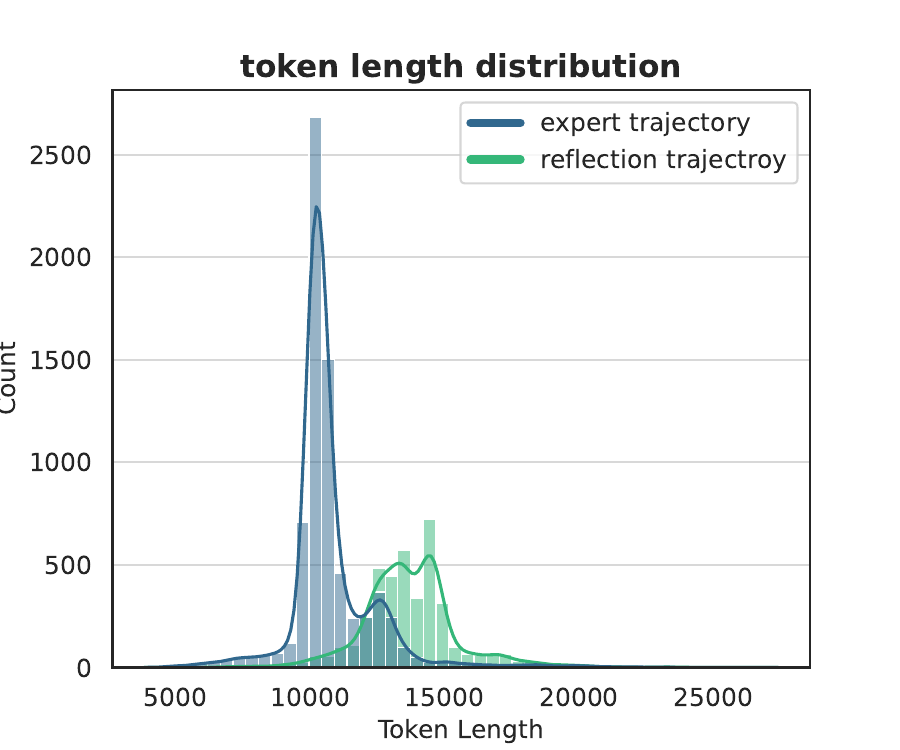}
        \label{fig:sub1}
    \end{subfigure}
    \begin{subfigure}[b]{0.46\textwidth}
        \includegraphics[width=\textwidth]{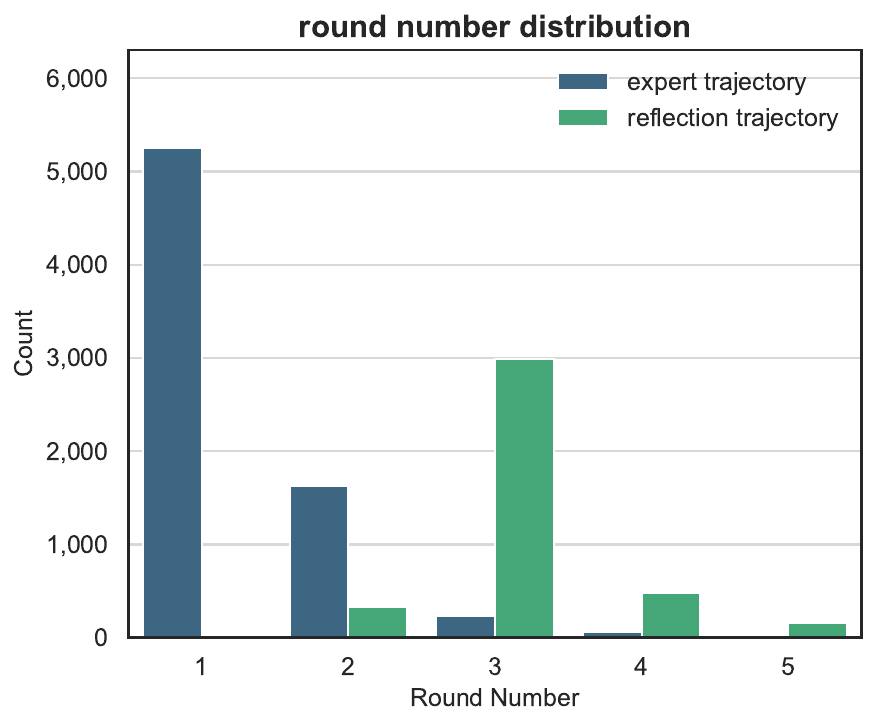}
        \label{fig:sub2}
    \end{subfigure}
    \vspace{-1.5em}
    \caption{Statistics of the cold start dataset.}
    \label{fig:stats}
\end{figure}
\begin{tcolorbox}[
    colback=gray!10!white,     
    colframe=black,            
    boxrule=1pt,               
    boxsep=4pt,                
    top=4pt, bottom=4pt,       
    left=8pt, right=8pt        
]
\begin{center}
\textbf{Reasoning Prompt}
\end{center}
\small
\textbf{System Prompt:} You are a helpful assistant. You will receive a low-frame-rate video and related questions. You can analyze the video content to answer the question and trigger high-frame-rate inspections when finer temporal resolution is needed. When you detect ambiguous motion/objects that require closer inspection, wrap your request in \texttt{<video\_zoom></video\_zoom>} tags and provide the exact time segment and target frame rate in JSON format: \texttt{<video\_zoom>\{"segment": [start\_sec, end\_sec], "fps": n\} </video\_zoom>}, it will return the video clip at the target fps to help you better answer the question. Note that the total frames num of the request clip cannot exceed 16 (e.g. (end\_sec - start\_sec) * fps $\leq$ 16) and DO NOT include \texttt{<answer>} tags in this round.  Example usage: \texttt{<video\_zoom> \{"segment": [4.0, 6.0], "fps": 2\} </video\_zoom>}.  If the initial tool response does not provide sufficient information to answer the question, you may continue to request additional video zoom inspections as needed, until you either (1) gather enough information to form a complete answer, or (2) are explicitly instructed to stop using the tool. Output the thinking process within \texttt{<think> </think>} tags, once you confirm your final answer place the final answer inside \texttt{<answer>} and \texttt{</answer>}.\\

\textbf{User:} ...\texttt{<framei\_timet><image>}...Question
\\
\end{tcolorbox}

\begin{tcolorbox}[
    breakable,
    colback=gray!10!white,     
    colframe=black,            
    boxrule=1pt,               
    boxsep=4pt,                
    top=4pt, bottom=4pt,       
    left=8pt, right=8pt        
]
\begin{center}
\textbf{Prompt for Constructing Reflection Trajectory}
\end{center}
\small
You are an expert video understanding model with access to a video zoom tool that allows you to request high-frame-rate clips for temporal inspection. Your task is to correct a flawed analysis of a low-frame-rate video by using a \texttt{video\_zoom} tool.
Your workflow is a multi-turn process:
\newline
\textbf{Turn 1: Reflection and Tool Call}
\begin{enumerate}
    \item \textbf{Analyze the Error}: You will be given a question, choices, and a previous, incorrect attempt. First, you must reflect on \textit{why} the previous \texttt{video\_zoom} tool call was flawed. Was the time segment wrong? Was the frames-per-second (fps) too low? Was the focus of the analysis misaligned with the question?
    \item \textbf{Formulate a Correction}: Based on your analysis, decide on a new, corrected \texttt{video\_zoom} request. This request should target the precise moment of interest and use an appropriate fps to capture the fine-grained detail.
    \item \textbf{Output the Tool Call}: Generate your reflection and the new tool call in the specified format. \textbf{Your output for this turn MUST end immediately after the \texttt{</video\_zoom>} tag.} Do not generate anything further. The system will then execute this call and provide you with the result.
\end{enumerate}
\textbf{Constraint for the tool call}: The total number of frames requested must not exceed 16. That is: \texttt{(end\_sec - start\_sec) * fps <= 16}.
\newline
\textbf{Turn 2: Analysis and Final Answer}
\begin{enumerate}
    \item \textbf{Receive Tool Response}: The system will provide the high-frame-rate video clip from your corrected tool call.
    \item \textbf{Analyze the New Clip}: Carefully examine the new clip. Describe what you can now clearly see that resolves the question.
    \item \textbf{Provide the Final Answer}: Based on your new observation, state the correct answer from the choices, enclosed in \texttt{\textbackslash boxed\{\}}.
\end{enumerate}
\textbf{Output Format Structure:}\newline
\textbf{[FIRST TURN OUTPUT]}
\newline
\texttt{<think>}
\newline
The previous tool call was incorrect because [explain the flaw in the tool use, e.g., wrong segment, wrong fps, or misaligned focus].
\newline
Now I will zoom in to inspect the motion of '\{target object/action\}' between \{start\_sec\}s and \{end\_sec\}s with higher temporal resolution.
\newline
\texttt{</think><video\_zoom> \{"segment": [start\_sec, end\_sec], "fps": n\} </video\_zoom>}
\newline
\textbf{[YOUR TURN 1 OUTPUT STOPS HERE]}\newline
\textbf{[SECOND TURN OUTPUT] (after you receive the tool response)}
\newline
\texttt{<think>}
In the corrected high-frame-rate clip, [describe what is clearly observed now].
\texttt{</think>}
\newline
\texttt{<answer>}
\texttt{\textbackslash\textbackslash boxed\{correct answer\}}
\texttt{</answer>}
\newline
\textbf{Example to follow:}\newline
\textbf{Question:} Which hand did the woman use to pick up the cup?\newline
\textbf{Choices:} A: Left hand B: Right hand C: Both hands D: Neither\newline
\textbf{Previous Trajectory (Wrong):}
Tool call: \texttt{<video\_zoom> \{"segment": [0.0, 2.0], "fps": 2\}</video\_zoom>}
\newline
\textbf{(Your First Turn Output Should Look Like This):}
\newline
\texttt{<think>}
The previous tool call was incorrect because it focused on the wrong time segment. The woman only reaches for the cup between 3.0s and 5.0s. Additionally, the low fps of 2 might not be sufficient to clearly distinguish the hand's motion.
\newline
Now I will zoom in to inspect the motion of 'the woman's hand reaching for the cup' between 3.0s and 5.0s with a higher temporal resolution.
\texttt{</think><video\_zoom> \{"segment": [3.0, 10.0], "fps": 1\} </video\_zoom>}
\newline
\textbf{(System provides tool response, then you start your Second Turn)}
\textbf{(Your Second Turn Output Should Look Like This):}
\newline
\texttt{<think>}In the corrected high-frame-rate clip, the woman's right hand is clearly seen moving towards and gripping the cup handle between 4.1s and 4.8s, while her left hand remains on her lap. The motion is now unambiguous. \texttt{</think>}
\texttt{<answer>} B. \texttt{</answer>}
\end{tcolorbox}
\section{More Experiment Results}
\subsection{Different Expert Model for Cold-Start Data Construction. }
In our main experiments, we utilized data distilled from Gemini 2.5 Pro to generate the cold-start SFT dataset. To justift this choice, we conduct a comparative analysis of data distilled from Gemini 2.5 pro versus data from GPT-4o. We conducted two identical training runs of our model. The only difference was the source of the cold start dataset used in the SFT stage: one model was trained on data distilled from GPT-4o, and the other on data from Gemini 2.5 Pro. Both models then underwent the same reinforcement learning phase. We evaluated the final performance of both models on our key benchmarks.
The results of this comparison are presented in Table \ref{tab:distillation_comparison}. As the result shows, the model trained using data from Gemini 2.5 pro  achieved slightly better results on most benchmarks. Through qualitative analysis of the generated data, we observed that the trajectories from Gemini exhibited greater diversity in their reasoning patterns and tool-use strategies.
\begin{table}[ht]
\centering
\caption{Performance comparison using different expert models for cold-start data construction.}
\label{tab:distillation_comparison}
\begin{adjustbox}{width=\linewidth,center}
\setlength{\tabcolsep}{1.5mm}
\renewcommand{\arraystretch}{1.5}
\begin{tabular}{lrccccccccc}
\toprule  
\multirow{2}{*}{\textbf{Model}} & \multicolumn{1}{c}{\multirow{2}{*}{\centering \textbf{Size}}} 
& \multicolumn{2}{c}{\textbf{MLVU}} & {\textbf{LongVideoBench } }  &\multicolumn{2}{c}{\textbf{VideoMME}} & {\textbf{LVBench}} & {\textbf{VideoMMLU}} & {\textbf{VideoMMMU}}  &{\textbf{LongVideoReason}}\\ 
& & \textbf{\textit{dev}} &  \textbf{\textit{test}}&\textbf{\textit{val}}&\textbf{\textit{overall}}&\textbf{\textit{long}}&  & \textbf{\textit{quiz} } & &\textbf{\textit{eval} }\\
\midrule
Qwen2.5-VL&7B   &58.1   &45.4   &51.0   &63.5   &53.9   &36.9   &61.0   &48.1   &70.8   \\
\ours$_{gemini}$ &7B   &68.8   &\textbf{55.8}   &\textbf{57.7}   &\textbf{65.2}   &\textbf{55.8}   &\textbf{41.5}   &\textbf{67.9}   &\textbf{52.2}   &\textbf{80.3}\\
\ours$_{gpt-4o}$ &7B   &\textbf{69.5}   &54.6   &55.5   &61.6   &51.0   &41.2   &64.1   &51.2   &78.4\\
\bottomrule
\end{tabular}
\end{adjustbox}
\end{table}

\subsection{\textcolor{blue}{Results on OOD tasks}} 
\textcolor{blue}{To assess the generalizability and robustness of \ours, we evaluated its performance on two distinct out-of-distribution (OOD) task categories: short video captioning and logical reasoning on synthetic data. These experiments were designed to verify that our training process enhances long-video capabilities without degrading the model's foundational abilities.}

\textcolor{blue}{While our primary focus is on long videos, we tested \ours on several short video captioning benchmarks TemporalBench\citep{cai2024temporalbench},TempCompass\citep{liu2024tempcompass} and VDC\citep{chai2024auroracap} to ensure its core descriptive capabilities were maintained. The results, summarized in Table \ref{tab:short_video_captioning}, show that our model not only preserves but significantly improves upon the baseline's performance across all tested benchmarks.}

\begin{table}[h!]
\centering
\caption{\textcolor{blue}{Short Video Captioning Benchmark Results}}
\label{tab:short_video_captioning}
\renewcommand{\arraystretch}{1.5}
\begin{tabular}{l|ccc}
\hline
 \multirow{2}{*}{\makecell[c]{\textbf{Model}}}& \textbf{TemporalBench} & \textbf{TempCompass} & \textbf{VDC} \\
 & \textbf{(Short Caption Score)} & \textbf{(Captioning Acc)} & \textbf{(Short Acc / Score)} \\
\hline
QwenVL-2.5-7B & 40.9 & 52.1 & 37.8 / 1.98 \\
\ours & 56.4 & 65.3 & 49.2 / 2.51 \\
\hline
\end{tabular}
\end{table}
\textcolor{blue}{We further tested the model's robustness on a subset of the CLEVRER dataset\citep{yi2019clevrer}, which evaluates causal and logical reasoning on synthetic videos. This domain is significantly different from the real-world, long-form videos used in our training.}

\textcolor{blue}{As shown in Table \ref{tab:clevrer_performance}, the comparable performance to the baseline model demonstrates that our two-stage training process does not degrade the model's foundational reasoning abilities. The minimal gain is expected, as the glance-and-zoom mechanism is not designed for the abstract, logical puzzles presented by CLEVRER. This result confirms that our method successfully retains the model's core competencies on tasks that do not require our agentic framework.}
\begin{table}[h!]
\centering
\caption{\textcolor{blue}{Performance on CLEVRER}}
\renewcommand{\arraystretch}{1.5}
\label{tab:clevrer_performance}
\begin{tabular}{l|c}
\hline
\textbf{Model} & \textbf{CLEVRER Accuracy} \\
\hline
QwenVL-2.5-7B  & 67.3 \\
\ours  & 68.0 \\
\hline
\end{tabular}
\end{table}

\subsection{\textcolor{blue}{Impact of SFT Data Quantity}}
 \textcolor{blue}{We investigated whether the effectiveness of our Supervised Fine-Tuning (SFT) phase stems from the quantity of data. We compared our model, trained on our curated ~11k trajectory dataset, against a model trained on a dataset of the same composition but with double the quantity ($\sim$20k samples).}

\begin{table}[h!]
\caption{\textcolor{blue}{Impact of SFT Data Quantity}}
\begin{adjustbox}{width=\linewidth,center}
\centering
\renewcommand{\arraystretch}{1.5}
\label{tab:sft_data_quantity}
\begin{tabular}{lccccc}
\toprule
\textbf{Training Dataset} & \textbf{MLVU (dev)} & \textbf{MLVU (test)} & \textbf{LVBench} & \textbf{LongVideoBench} & \textbf{LongVideoReason-eval} \\
\midrule
Ours ($\sim$11k) & 68.8 & 55.8 & 41.5 & 57.7 & 80.3 \\
Scaled Dataset ($\sim$20k) &66.4 & 56.0 & 41.4 & 55.6 & 80.3 \\
\bottomrule
\end{tabular}
\end{adjustbox}
\end{table}
\textcolor{blue}{As shown in Table \ref{tab:sft_data_quantity}, simply doubling the data quantity did not lead to better overall performance. While there was a marginal improvement on MLVU (dev), the model trained on the larger dataset performed worse on all other benchmarks. This result strongly suggests that the effectiveness of our dataset comes from the high-quality, diverse reasoning patterns it contains, rather than its sheer size. This \enquote{less is more} philosophy aligns with recent findings from works like DeepSeek-R1\citep{guo2025deepseek} and LIMO\citep{ye2025limo}, which demonstrate that a few thousand high-quality, reasoning-focused samples can be sufficient to unlock powerful capabilities in large models. Our methodology prioritizes a rich collection of reasoning pathways over a large volume of repetitive examples.}
\subsection{\textcolor{blue}{Analysis of Chosen FPS}}
\textcolor{blue}{A key feature of our \texttt{<video\_zoom>} tool is that the frames-per-second (fps) for a \enquote{zoom-in} clip is dynamically generated by the model itself, allowing it to decide not only where to look but also how closely to look. To understand the model's learned behavior, we analyzed the distribution of fps values it chose across thousands of tool calls on our validation set.}

\textcolor{blue}{The results in Table \ref{tab:fps_distribution} reveal that the model does not default to the highest possible fps. Instead, its most frequent choice is a moderate fps in the (1, 2] range, which it selects in 66.2\% of cases. This demonstrates that the model learns an efficient policy, requesting just enough temporal detail to solve the task without unnecessarily expending its frame budget. While a high fps like 8 might seem excessive for a full video, it is a reasonable and effective choice for examining a critical few-second clip, and the model learns to use it sparingly.}
\begin{table}[h!]
\centering
\renewcommand{\arraystretch}{1.5}
\caption{\textcolor{blue}{Distribution of \texttt{fps} Values Chosen by the Model}}
\label{tab:fps_distribution}
\begin{tabular}{lccccc}
\toprule
\textbf{fps Range} & \textbf{(0, 1]} & \textbf{(1, 2]} & \textbf{(2, 4]} & \textbf{(4, 8]} & \textbf{(8, \(\infty\))} \\
\midrule
\textbf{Percentage} & 24.9\% & 66.2\% & 8.2\% & 0.6\% & $\le$0.1\% \\
\bottomrule
\end{tabular}
\end{table}

\subsection{\textcolor{blue}{Performance by Maximum Allowed Tool Calls}}
\textcolor{blue}{To understand the impact of multi-step reasoning, we evaluated how model accuracy changes with the maximum number of allowed tool calls. We varied the limit from 0 (no tool use) to 4 and measured performance across several benchmarks.}

\begin{table}[h!]
\caption{\textcolor{blue}{Performance by Maximum Allowed Tool Calls}}
\centering
\begin{adjustbox}{width=\linewidth,center}
\renewcommand{\arraystretch}{1.5}
\label{tab:perf_by_tool_calls}
\begin{tabular}{lccccc}
\toprule
\textbf{Max Tool Calls} & \textbf{MLVU (dev)} & \textbf{MLVU (test)} & \textbf{LongVideoBench} & \textbf{LVBench} & \textbf{LongVideoReason-eval} \\
\midrule
0 (No Tool Use) & 65.0 & 52.6 & 52.7 & 38.7 & 64.8 \\
1 & 66.7 & 53.4 & 55.3 & 39.6 & 79.9 \\
2 & 67.8 & 56.6 & 57.0 & 41.4 & 80.2 \\
3 & 68.4 & 56.0 & 56.2 & 41.8 & 81.1 \\
4 & 68.8 & 55.8 & 57.7 & 41.5 & 80.3 \\
\bottomrule
\end{tabular}
\end{adjustbox}
\end{table}
\textcolor{blue}{The results in Table \ref{tab:perf_by_tool_calls} show a clear trend. The most significant performance jump occurs when allowing just one tool call, confirming the fundamental value of our agentic framework. For example, on LongVideoReason-eval, accuracy leaps from 64.8\% to 79.9\%.
Increasing the limit from one to two calls provides another consistent and significant boost, highlighting the importance of multi-turn interaction for refining searches or gathering more evidence. After two calls, the benefits of increasing the limit further show diminishing returns. While there are minor gains, the improvements are smaller and less consistent. Our final model, with a limit of 4, uses this extra budget sparingly for the hard problems, but the core performance is achieved within the first two interactions.
}

\section{More Cases}
\begin{figure}[htb]
    \centering
    \includegraphics[width=0.9\linewidth]{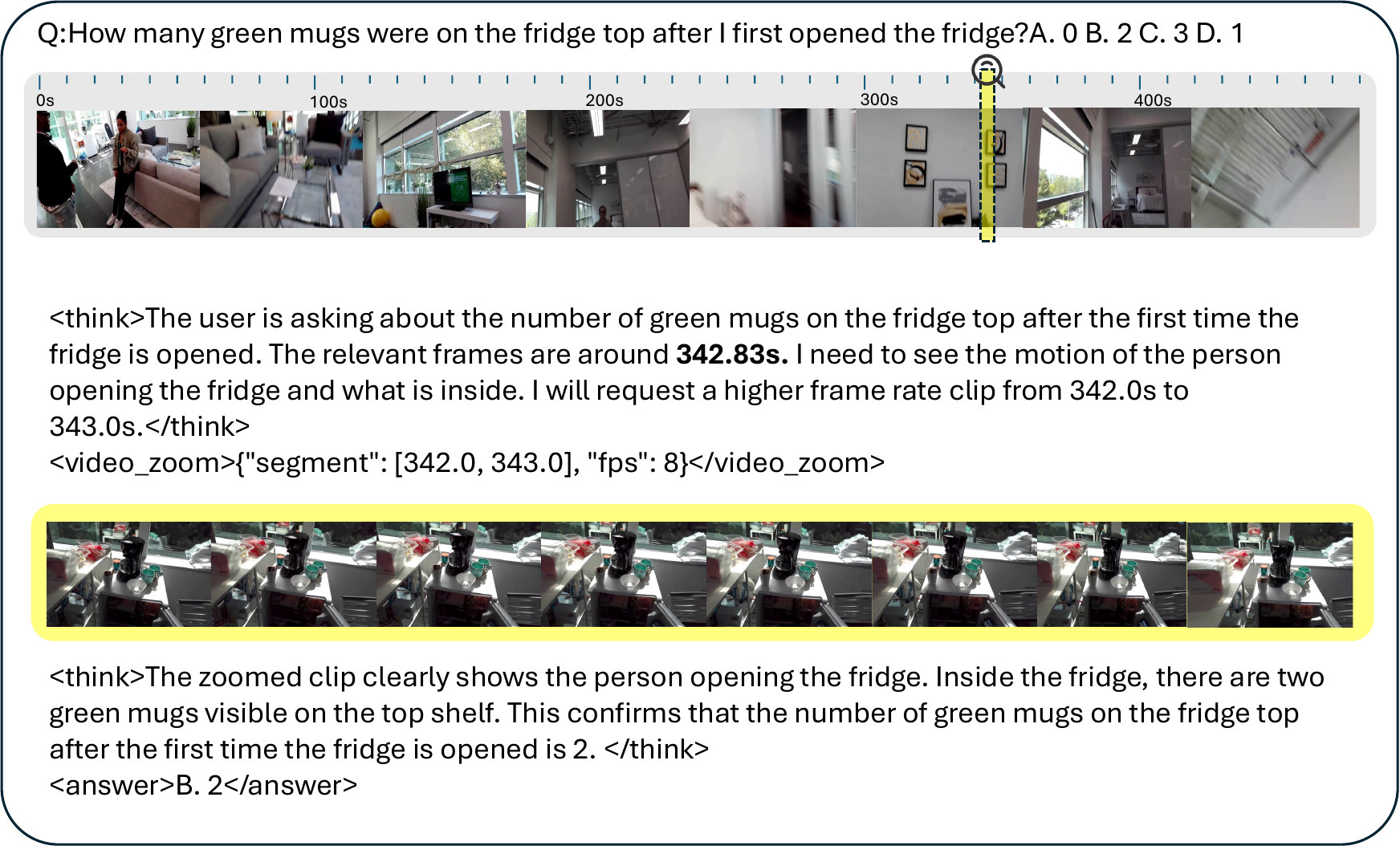}
    \caption{
        Case Study: Single-Detail Reasoning.
    }
    \label{fig:case1}
\end{figure}

\begin{figure}[htb]
    \centering
    \includegraphics[width=0.9\linewidth]{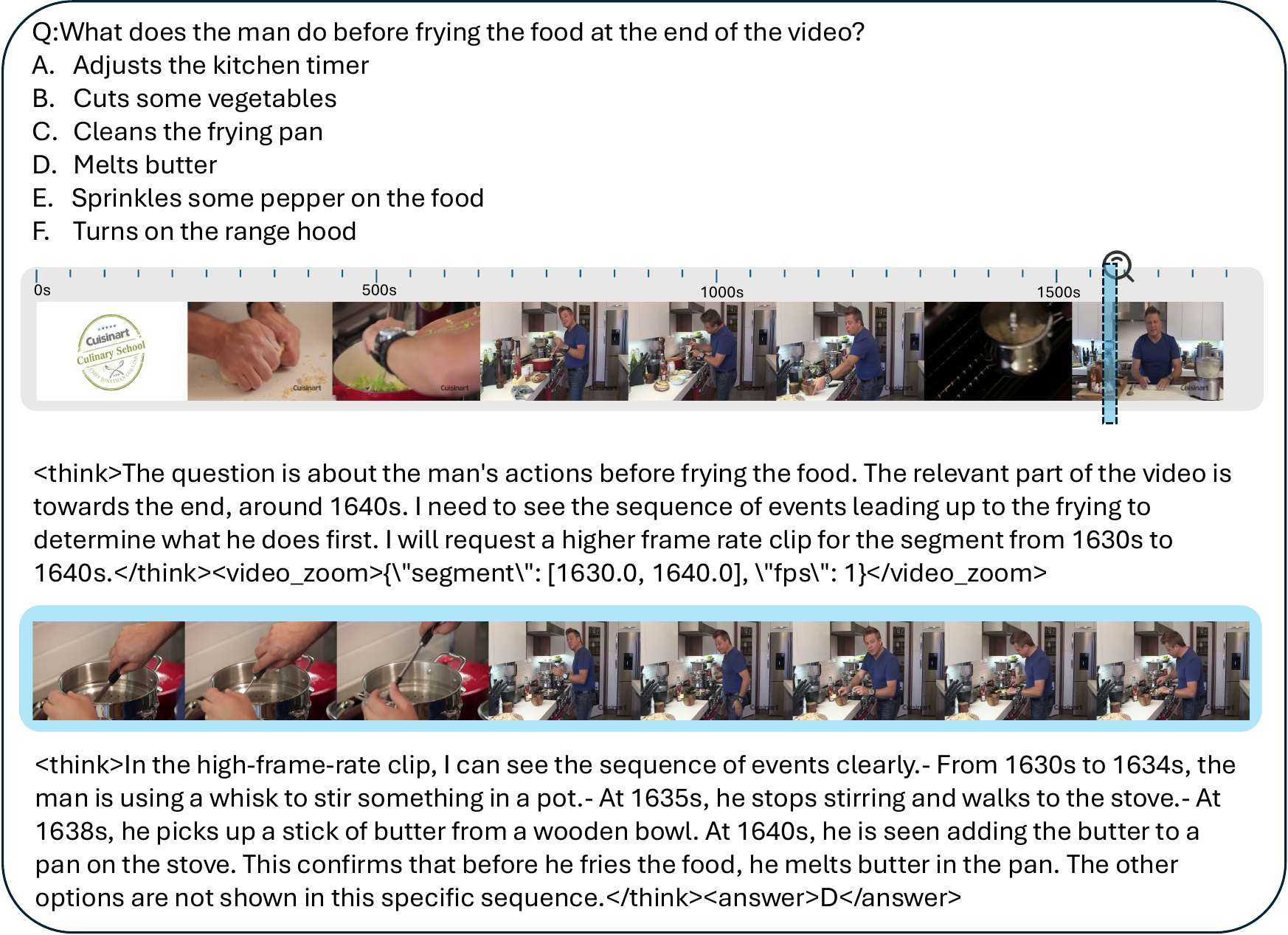}
    \caption{
        \textcolor{blue}{Case Study: Single-Detail Reasoning.}
    }
    \label{fig:case1}
\end{figure}

\begin{figure}[htb]
    \centering
    \includegraphics[width=0.9\linewidth]{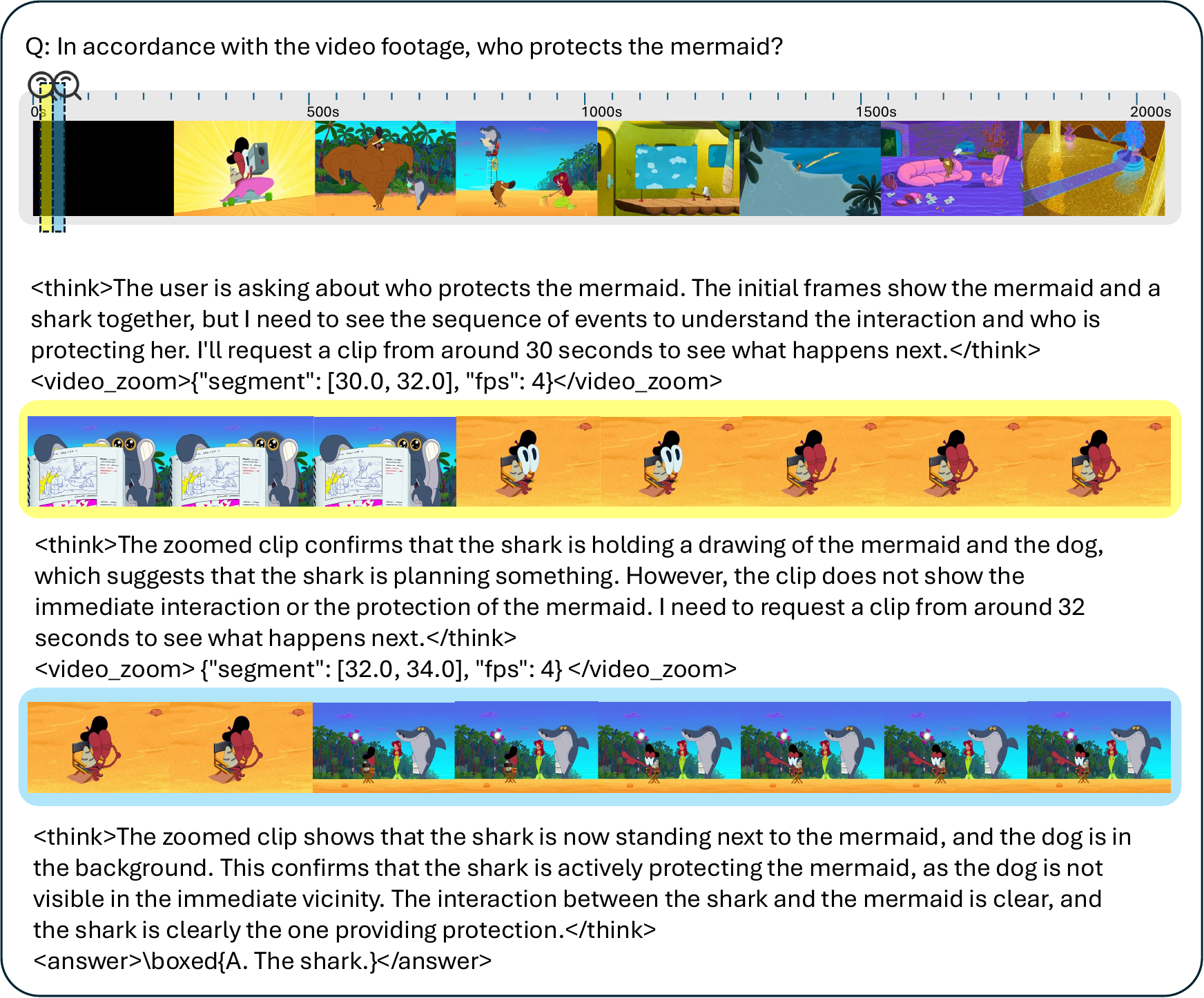}
    \caption{
        Case Study: Self-Correction and Refined Reasoning.
    }
    \label{fig:case2}
\end{figure}

\begin{figure}[htb]
    \centering
    \includegraphics[width=0.9\linewidth]{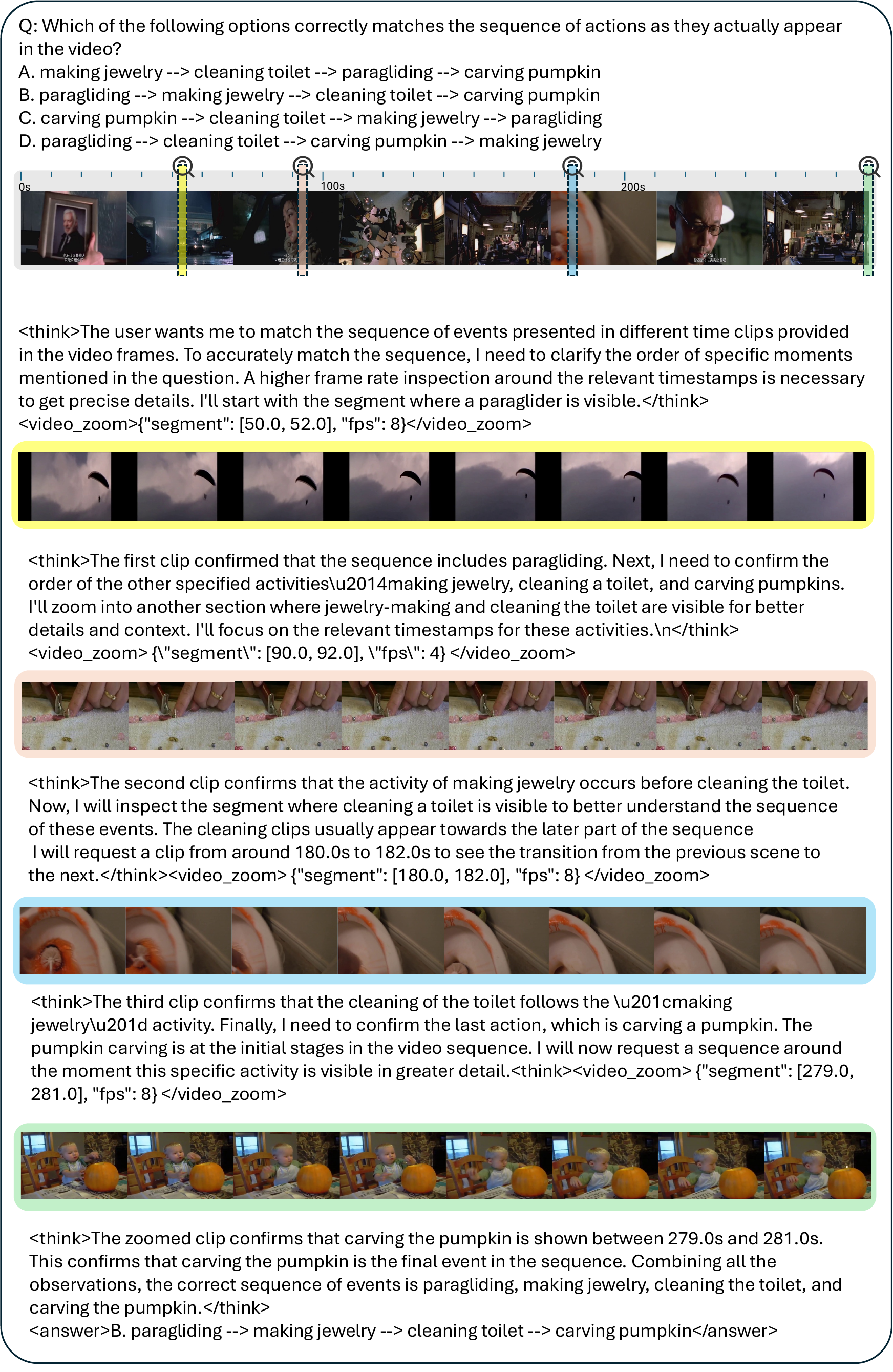}
    \caption{
        Case Study: Multi-Detail Reasoning for Complex Queries.
    }
    \label{fig:case3}
\end{figure}

\begin{figure}[h]
    \centering
    \includegraphics[width=1\linewidth]{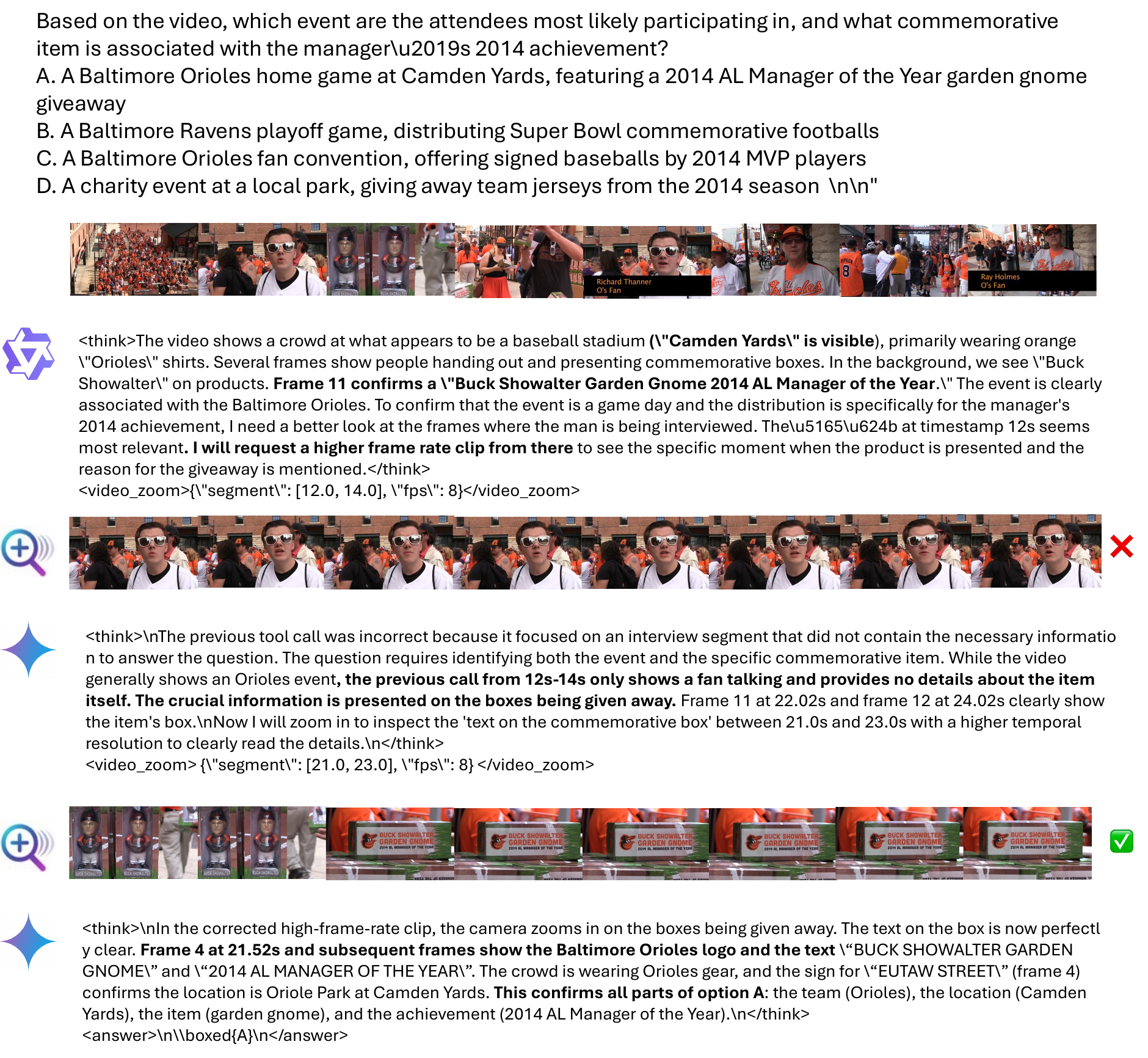}
    \caption{Demostration of cold start data.}
\end{figure}
\begin{figure}[h]
    \centering
    \includegraphics[width=1\linewidth]{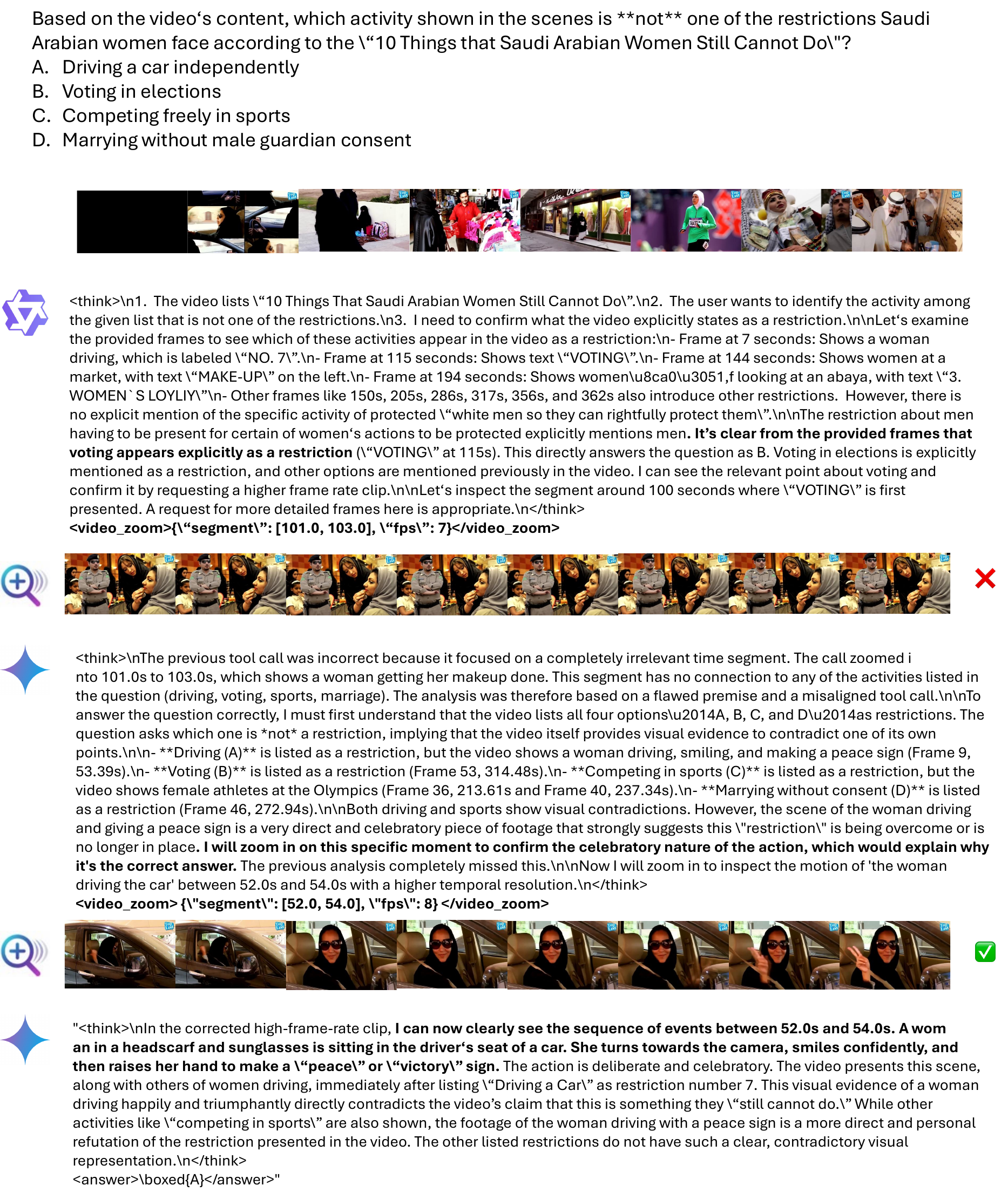}
    \caption{Example of cold start data.}
\end{figure}

\end{document}